\documentclass[runningheads]{llncs}

\usepackage{eccv}

\usepackage{eccvabbrv}

\usepackage{graphicx}
\usepackage{booktabs}

\usepackage[accsupp]{axessibility}  %

\usepackage{graphicx, amsmath, amssymb, caption, subcaption, multirow, overpic, textpos, multibib}

\usepackage{graphicx}
\usepackage{tikz}
\usepackage{comment}
\usepackage{amsmath,amssymb} %
\usepackage{color}
\usepackage{enumitem} 

\usepackage{wrapfig}
\usepackage{lipsum}
\usepackage{comment}
\usepackage{algorithm}
\usepackage{algorithmic}
\usepackage{listings}
\usepackage{wrapfig}
\usepackage{subcaption}

\usepackage[british,english,american]{babel}
\definecolor{citecolor}{HTML}{0071BC}
\definecolor{linkcolor}{HTML}{ED1C24}
\newcommand{\app}{\raise.17ex\hbox{$\scriptstyle\sim$}}

\usepackage{xspace}
\makeatletter
\DeclareRobustCommand\onedot{\futurelet\@let@token\@onedot}
\def\@onedot{\ifx\@let@token.\else.\null\fi\xspace}

\makeatother

\usepackage{comment}

\newlength\savewidth\newcommand\shline{\noalign{\global\savewidth\arrayrulewidth
  \global\arrayrulewidth 1pt}\hline\noalign{\global\arrayrulewidth\savewidth}}

\newcommand{\tablestyle}[2]{\ttfamily\setlength{\tabcolsep}{#1}\renewcommand{\arraystretch}{#2}\centering\footnotesize}

\definecolor{gain}{HTML}{34a853}  %

\definecolor{lost}{HTML}{ea4335}  %

\newcites{app}{Appendix References}

\makeatletter\renewcommand\paragraph{\@startsection{paragraph}{4}{\z@}
  {.5em \@plus1ex \@minus.2ex}{-.5em}{\normalfont\normalsize\bfseries}}\makeatother

\definecolor{baselinecolor}{gray}{.9}

\usepackage[dvipsnames]{xcolor}
\usepackage{colortbl}

\newcommand{\E}{\mathbb{E}}

\newcommand{\LL}{\mathcal{L}}

\newcommand{\rd}{{\mathrm{d}}} 
\newcommand{\bx}{{\mathbf{x}}}
\newcommand{\bX}{{\mathbf{X}}}

\newcommand{\bv}{{\mathbf{v}}}
\newcommand{\bs}{{\mathbf{s}}}
\newcommand{\bS}{{\mathbf{S}}}
\newcommand{\bW}{{\mathbf{W}}}

\newcommand{\bz}{{\mathbf{z}}}
\newcommand{\eps}{{\boldsymbol \varepsilon}}

\definecolor{ForestGreen}{RGB}{34,139,34}

\usepackage[pagebackref,breaklinks,colorlinks,citecolor=eccvblue]{hyperref}

\usepackage{orcidlink}

\definecolor{mycolor}{HTML}{5a860e}
\def\logo{\makebox[22pt][l]{\raisebox{-0.9ex}{\includegraphics[height=30pt]{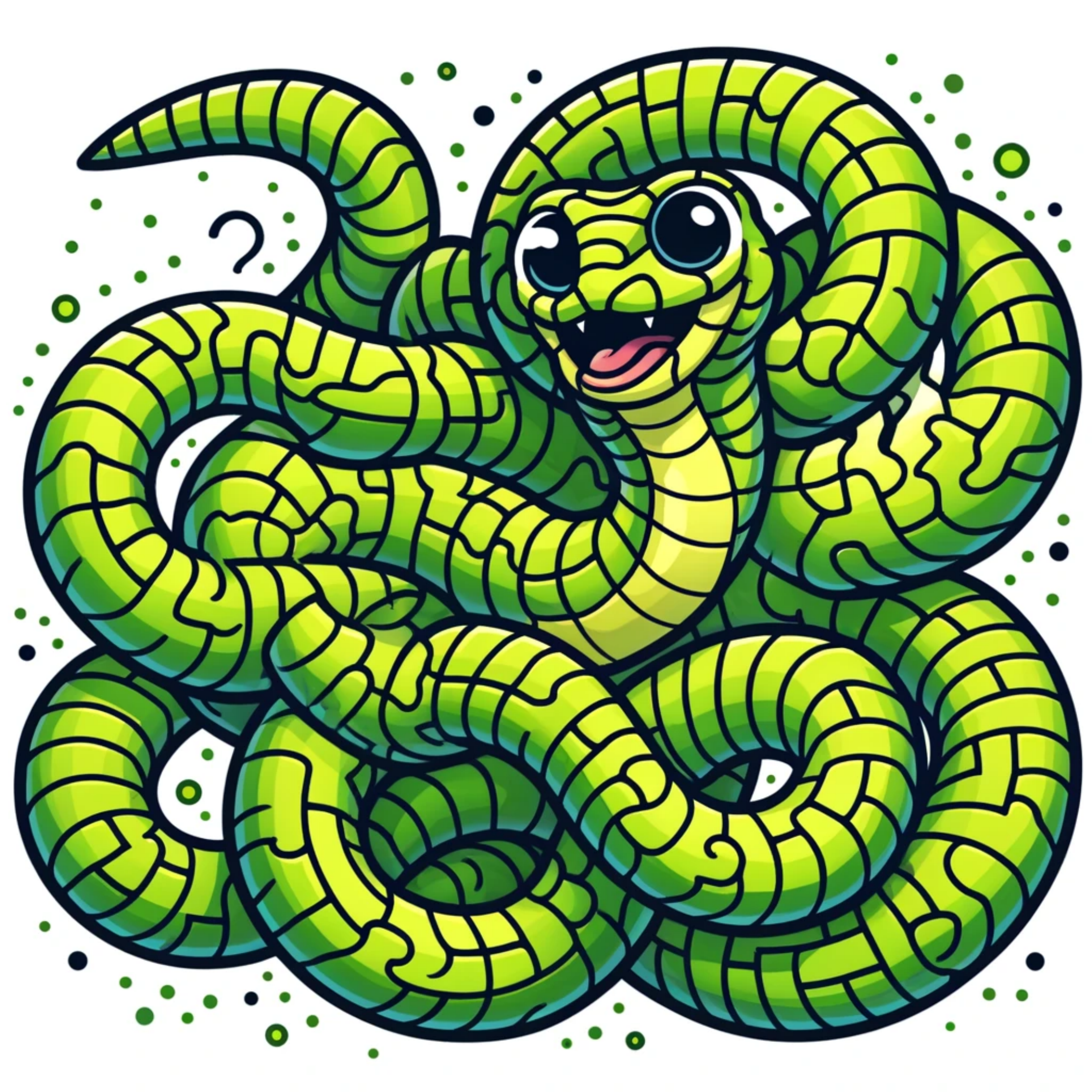}}}\hspace{10pt}}

\begin{document}

\title{\logo\textcolor{mycolor}{ZigMa}: A DiT-style \textcolor{mycolor}{Zig}zag \textcolor{mycolor}{Ma}mba Diffusion Model} 

\titlerunning{ZigMa}

\author{Vincent Tao Hu\and
Stefan Andreas Baumann\and
Ming Gui  \and \\
Olga Grebenkova \and  Pingchuan Ma \and Johannes Schusterbauer \and Björn Ommer}

\authorrunning{Hu et al.}

\institute{
CompVis @ LMU Munich, MCML
\\
\url{https://compvis.github.io/zigma/} }

\maketitle

\begin{abstract}
  The diffusion model has long been plagued by scalability and quadratic complexity issues, especially within transformer-based structures. In this study, we aim to leverage the long sequence modeling capability of a  State-Space Model called Mamba to extend its applicability to visual data generation. Firstly, we identify a critical oversight in most current Mamba-based vision methods, namely the lack of consideration for spatial continuity in the scan scheme of Mamba. Secondly, building upon this insight, we introduce Zigzag Mamba, a simple, plug-and-play,  minimal-parameter burden, DiT style solution, which outperforms Mamba-based baselines and demonstrates improved speed and memory utilization compared to transformer-based baselines, also this heterogeneous layerwise scan enables zero memory and speed burden when we consider more scan paths.
Lastly, we integrate Zigzag Mamba with the Stochastic Interpolant framework to investigate the scalability of the model on large-resolution visual datasets, such as FacesHQ $1024\times 1024$ and UCF101, MultiModal-CelebA-HQ, and MS COCO $256\times 256$.


\end{abstract}

\section{Introduction}

Diffusion models have demonstrated significant advancements across various applications, including image processing~\cite{sgdm,sgfm,rombach2022high_latentdiffusion_ldm}, video analysis~\cite{ho2022video}, point cloud processing~\cite{wu2023fast}, representation learning~\cite{fuest2024diffusion} and human pose estimation~\cite{gong2023diffpose}. Many of these models are built upon Latent Diffusion Models (LDM)\cite{rombach2022high_latentdiffusion_ldm}, which are typically based on the UNet backbone. However, scalability remains a significant challenge in LDMs\cite{huang2024scalelong}. Recently, transformer-based structures have gained popularity due to their scalability~\cite{dit_peebles2022scalable,uvit} and effectiveness in multi-modal training~\cite{bao2023one}. Notably, the transformer-based structure DiT~\cite{dit_peebles2022scalable} has even contributed to enhancing the high-fidelity video generation model SORA~\cite{sora} by OpenAI. Despite efforts to alleviate the quadratic complexity of the attention mechanism through techniques such as windowing~\cite{swin_transformer}, sliding~\cite{beltagy2020longformer}, sparsification~\cite{child2019generating,kitaev2020reformer}, hashing~\cite{choromanski2020rethinking,sun2021sparse}, Ring Attention~\cite{liu2023ring,brandon2023striped}, Flash Attention~\cite{dao2022flashattention} or a combination of them~\cite{ao2024burstattention,ring_flash_attention}, it remains a bottleneck for diffusion models.

\begin{figure}
    \centering
    \includegraphics[width=0.7\textwidth]{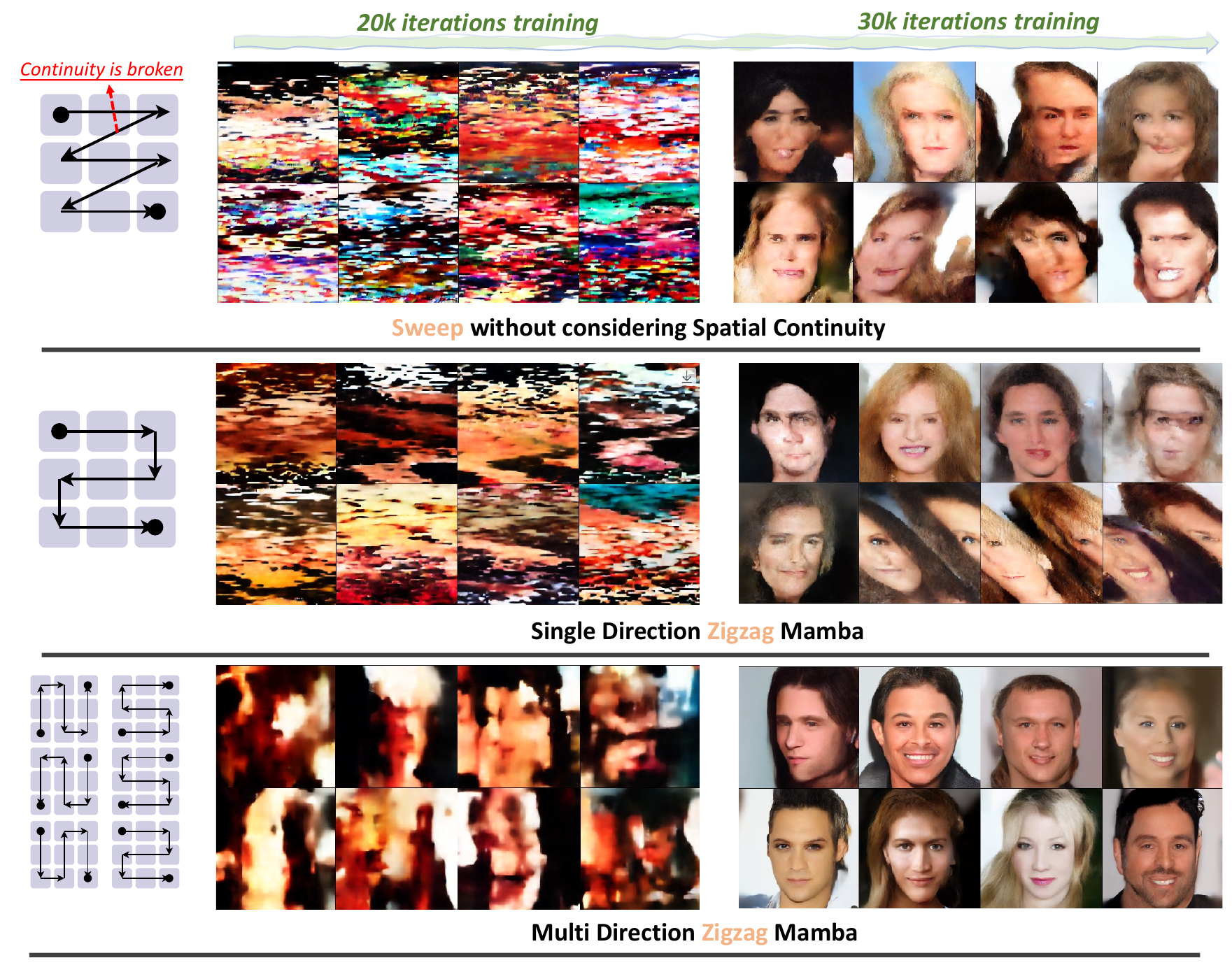}
    \caption{\textbf{Motivation.} Our Zigzag Mamba method improves the network's position-awareness by arranging and rearranging the scan path of Mamba in a heuristic manner. }
     \label{fig:teaser}
\end{figure}

On the other hand, State-Space Models~\cite{gu2021efficiently,gupta2022diagonal,gu2022parameterization} have demonstrated significant potential for long sequence modeling, rivaling transformer-based methods. Their biological similarity~\cite{tikochinski2024incremental} and efficient memory state also advocate for the use of the State-Space model over the transformer.
Several methods~\cite{gu2023mamba,gu2021efficiently,fu2022hungry_h3,smith2022simplified_s4} have been proposed to enhance the robustness~\cite{yu2023robustifying}, scalability~\cite{gu2023mamba}, and efficiency~\cite{gu2021efficiently,gu2021combining} of State-Space Models. 
Among these, a method called Mamba~\cite{gu2023mamba} aims to alleviate these issues through work-efficient parallel scanning and other data-dependent innovations. However, the advantage of Mamba lies in 1D sequence modeling, and extending it to 2D images is a challenging question. Previous works~\cite{visionmamba,liu2024vmamba} have proposed flattening 2D tokens directly by computer hierarchy such as  row-and-column-major order, but this approach neglects \emph{Spatial Continuity}, as shown in Figure~\ref{fig:teaser}. Other works~\cite{liu2024swinmamba,umamba} consider various directions in a single Mamba block, but this introduces additional parameters and GPU memory burden. In this paper, we aim to emphasize the importance of \emph{Spatial Continuity} in Mamba and propose several intuitive and simple methods to enable the application of Mamba blocks to 2D images by incorporating continuity-based inductive biases in images. We also generalize these methods to 3D with spatial-temporal factorization on 3D sequence.

In the end, Stochastic Interpolant~\cite{albergo2023stochastic} provides a more generalized framework  that can uniform various generative models including, Normalizing Flow~\cite{neuralode_chen2018neural}, diffusion model~\cite{sohl2015deep,ho2020denoising,song2021scorebased_sde}, Flow matching~\cite{lipman2022flow,rectifiedflow_iclr23,albergo2022building}, and Schrödinger Bridge~\cite{liu2022deep_gdsb}. Previously, some works~\cite{ma2024sit} explore the Stochastic Interpolant on relatively small resolutions, e.g., $256\times 256$, $512\times 512$. In this work, we aim to explore it in further more complex scenarios e.g., $1024 \times 1024$ resolution and even in videos.

In summary, our contributions are as follows: Firstly, we identify the critical issue of \emph{Spatial Continuity} in generalizing the Mamba block from 1D sequence modeling to 2D image and 3D video modeling. Building on this insight, we propose a simple, plug-and-play, zero-parameter heterogeneous layerwise scan paradigm named \emph{Zigzag Mamba (ZigMa)} that leverages spatial continuity to maximally incorporate the inductive bias from visual data. %
Secondly, we extend the methodology from 2D to 3D by factorizing the spatial and temporal sequences to optimize performance. Secondly, we provide comprehensive analysis surrounding the Mamba block within the regime of diffusion models. Lastly, we demonstrate that our designed \emph{Zigzag Mamba} outperforms related Mamba-based baselines, representing the first exploration of Stochastic Interpolants on large-scale image data ($1024 \times 1024$) and videos.

\section{Related Works}

\noindent\textbf{Mamba.} Several works~\cite{wang2024state,wang2023stablessm,wang2024state} have demonstrated that the State-Space Model possesses universal approximation ability under certain conditions. Mamba, as a new State-Space Model, has superior potential for modeling long sequences efficiently, which has been explored in various fields such as medical imaging~\cite{xing2024segmamba,umamba,wang2024mamba,ruan2024vm}, video~\cite{li2024videomamba,park2024videomamba}, image restoration~\cite{guo2024mambair,zheng2024u_mamba_dehazing}, graphs~\cite{behrouz2024graphmamba}, NLP word byte~\cite{wang2024mambabyte}, tabular data~\cite{ahamed2024mambatab}, point clouds~\cite{liang2024pointmamba}, human motion~\cite{zhang2024motionmamba,wang2024text}, multi-task~\cite{lin2024mtmamba} and image generation~\cite{fei2024scalable_dis}.  Among them, the most related to us are VisionMamba\cite{visionmamba,liu2024vmamba}, S4ND~\cite{nguyen2022s4nd} and Mamba-ND~\cite{li2024mamba_nd}.  VisionMamba\cite{visionmamba,liu2024vmamba} uses a bidirectional SSM in discriminative tasks which incurs a high computational cost. Our method applies a simple alternative mamba diffusion in generative models.
S4ND~\cite{nguyen2022s4nd} introduces local convolution into Mamba's reasoning process, moving beyond the use of only 1D data.  Mamba-ND~\cite{li2024mamba_nd} takes multi-dimensionality into account in discriminative tasks, making use of various scans within a single block. In contrast, our focus is on distributing scan complexity across every layer of the network, thus maximizing the incorporation of inductive bias from visual data with zero parameter burden.
Scan curve is an important direction in SSM, PointMamba~\cite{liang2024pointmamba} is a representative work that employs SSM with space curves (e.g., Hilbert) for point cloud analysis, achieving remarkable performance. In contrast with them, our preliminary results show that the Hilbert curve doesn't work well with our method (see Appendix), while our method can be regarded as the simplest Peano curve.
For more information related to Mamba's work, please refer to the survey~\cite{wang2024state_survey}.

\noindent\textbf{Backbones in Diffusion Models.} Diffusion models primarily employ UNet-based~\cite{ho2020denoising,rombach2022high_latentdiffusion_ldm} and ViT-based~\cite{dit_peebles2022scalable, uvit} backbones. While UNet is known for high memory demands~\cite{rombach2022high_latentdiffusion_ldm}, ViT benefits from scalability~\cite{dehghani2023scaling,chen2023gentron} and multi-modal learning~\cite{bao2023one}. However, ViT's quadratic complexity limits visual token processing, prompting studies towards mitigating this issue~\cite{1024words,dao2022flashattention,beltagy2020longformer}. Our work, inspired by Mamba~\cite{gu2023mamba}, explores an SSM-based model as a generic diffusion backbone, retaining ViT's modality-agnostic and sequential modeling advantages. Concurrently, DiffSSM~\cite{yan2023_diffusionwithoutattention} concentrates on unconditional and class conditioning within the S4 model~\cite{gu2021efficiently}. DIS~\cite{fei2024scalable_dis} mainly explores the state-space model on a relatively small resolution, which is not the exact focus of our work. 
Our work significantly differs from theirs as it primarily focuses on the backbone design using the Mamba block and extends it to text conditioning. Furthermore, we apply our method to more complex visual data.

\noindent\textbf{SDE and ODE in Diffusion models.} The realm of Score-based Generative Models encompasses significant contributions from foundational works such as Score Matching with Langevin Dynamics (SMLD) by Song et al.\cite{song2019generative}, and the advent of Diffusion Models with Denoising Score Matching (DDPMs) proposed by Ho et al.\cite{ho2020denoising}. These methodologies operate within the framework of Stochastic Differential Equations (SDEs), a concept further refined in the research of Song et al.~\cite{song2021scorebased_sde}.
Recent research strides, as exemplified by Karras et al.\cite{karras2022elucidating} and Lee et al.\cite{lee2023minimizing}, have showcased the efficacy of employing Ordinary Differential Equation (ODE) samplers for diffusion SDEs, offering significant reductions in sampling costs compared to traditional approaches that entail discretizing diffusion SDEs. Furthermore, within the domain of Flow Matching~\cite{lipman2022flow} and Rectified Flow~\cite{liu2022flow}, both SMLD and DDPMs emerge as specialized instances under distinct paths of the Probability Flow ODE framework~\cite{song2021scorebased_sde}, with broad applications in vision~\cite{hulfm,schusterbauer2023boosting,dao2023flow_lfm}, depth~\cite{gui2024depthfm}, human motion~\cite{motionfm}, even language~\cite{flowseq}. These models typically utilize velocity field parameterizations employing the linear interpolant, a concept that finds broader applications in the Stochastic Interpolant framework~\cite{albergo2023stochastic}, with subsequent generalizations extending to manifold settings~\cite{benhamu2022}.
The SiT model~\cite{ma2024sit} scrutinizes the interplay between interpolation methods in both sampling and training contexts, albeit in the context of smaller resolutions such as $512\times 512$. Our research endeavors to extend these insights to a larger scale, focusing on the generalization capabilities for 2D images of $1024 \times 1024$ and 3D video data.

\section{Method}

In this section, we begin by providing background information on State-Space Models~\cite{gu2021efficiently,gupta2022diagonal,gu2022parameterization}, with a particular focus on a special case known as Mamba~\cite{gu2023mamba}. We then highlight the critical issue of \emph{Spatial Continuity} within the Mamba framework, and based on this insight, we propose the Zigzag Mamba. This enhancement aims to improve the efficiency of 2D data modeling by incorporating the continuity inductive bias inherent in 2D data. Furthermore, we design a basic cross-attention block upon Mamba block to achieve text-conditioning. Subsequently, we suggest extending this approach to 3D video data by factorizing the model into spatial and temporal dimensions, thereby facilitating the modeling process. Finally, we introduce the theoretical aspects of stochastic interpolation for training and sampling, which underpin our network architecture.

\subsection{Background: State-Space Models}
State Space Models (SSMs)~\cite{gu2021efficiently,gupta2022diagonal,gu2022parameterization} have been proven to handle long-range dependencies theoretically and empirically~\cite{gu2021combining} with linear scaling w.r.t sequence length. In their general form, a linear state space model can be written as follows:
\begin{equation*}
\begin{aligned}
    x'(t) &= \mathbf{A}(t)x(t) + \mathbf{B}(t)u(t) \\
    y(t) &= \mathbf{C}(t)x(t) + \mathbf{D}(t)u(t),
\end{aligned}
    \label{eq:ssm}
\end{equation*}
mapping a 1-D input sequence $u(t) \in \mathbb{R}$ to a 1-D output sequence $y(t) \in \mathbb{R}$ through an implicit N-D latent state sequence $x(t) \in \mathbb{R}^n$. Concretely, deep SSMs seek to use stacks of this simple model in a neural sequence modeling architecture, where the parameters $\mathbf{A}, \textbf{B}, \mathbf{C}$ and $\mathbf{D}$ for each layer can be learned via gradient descent.

\begin{figure*}
    \centering
    \includegraphics[width=0.8\textwidth]{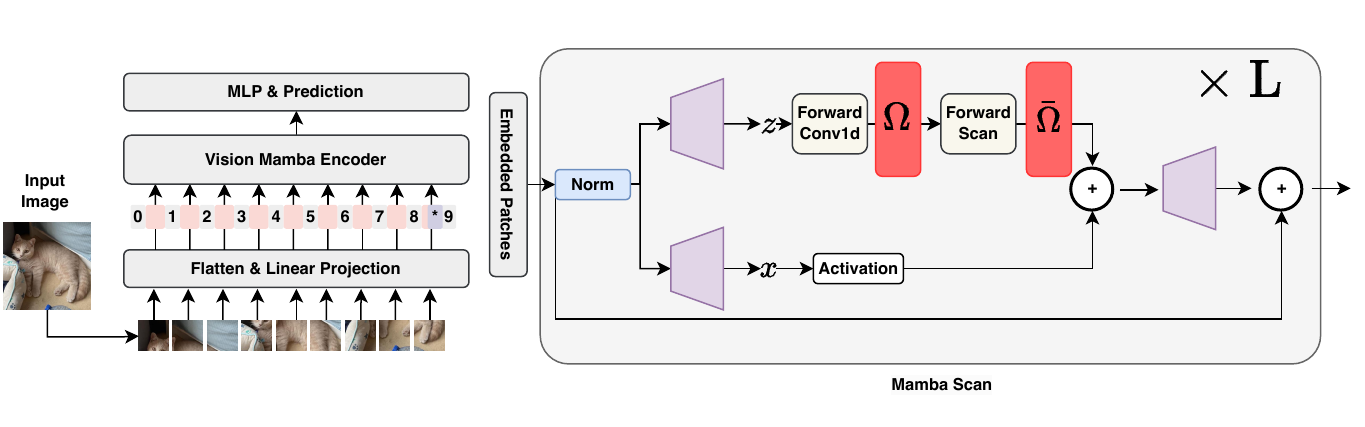}
    \caption{\textbf{ZigMa.} Our backbone is structured in L layers, mirroring the style of DiT~\cite{dit_peebles2022scalable}. We use the single-scan Mamba block as the primary reasoning module across different patches. To ensure the network is positionally aware, we've designed an arrange-rearrange scheme based on the single-scan Mamba. Different layers follow pairs of unique rearrange operation \textcolor{red}{$\Omega$} and reverse rearrange \textcolor{red}{$\Bar{\Omega}$}, optimizing the position-awareness of the method.
    }
     \label{fig:framework}
\end{figure*}

Recently, Mamba~\cite{gu2023mamba} largely improved the flexibility of SSMs in Language Modelling by relaxing the time-invariance constraint on SSM parameters, while maintaining computational efficiency. 
Several studies~\cite{visionmamba,liu2024vmamba} have been conducted to adapt the use of Mamba from unidimensional language data to multidimensional visual data. While most of these studies try to duplicate the $\textbf{A}$ to facilitate the new (reversed) direction, this approach can lead to additional parameters and an increased memory burden.
In this paper, we focus on exploring the scanning scheme of Mamba in diffusion models to efficiently maximize the use of inductive-bias from multi-dimensional visual data with zero parameter and memory burden.

\subsection{Diffusion Backbone: Zigzag Mamba}

\begin{figure*}
    \centering
\includegraphics[width=0.65\textwidth]{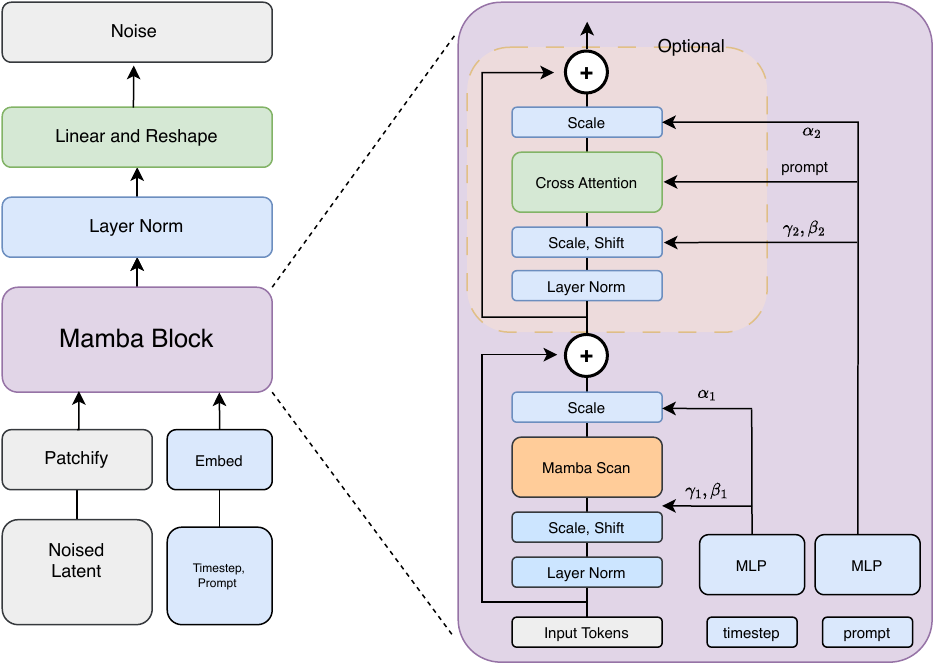}
    \caption{\textbf{The Detail of our Zigzag Mamba block. 
    } The detail of Mamba Scan is shown in Figure~\ref{fig:framework}. The condition can include a timestep and a text prompt. These are fed into an MLP, which separately modulates the Mamba scan for long sequence modeling and cross-attention for multi-modal reasoning.
    }
\label{fig:block_zigzagmamba}
\end{figure*}

\textbf{DiT-Style Network.} We opt to use the framework of DiT by AdaLN~\cite{dit_peebles2022scalable} rather than the skip-layer focused U-ViT structure~\cite{uvit}, as DiT has been validated as a scalable structure in literature~\cite{sora,bao2023one,chen2023gentron}.
Additionally, the Hourglass structure with downsampling~\cite{hourglass,unet} requires selecting the depth and width based on the complexity of the dataset and task. This requirement limits the flexibility of the solution.
Considering the aforementioned points, it informs our Mamba network design depicted in Figure~\ref{fig:block_zigzagmamba}. 
The core component of this design is the Zigzag Scanning, which will be explained in the following paragraph.

\noindent\textbf{Zigzag Scanning in Mamba.} Previous studies~\cite{bigs_wang2022pretraining,yan2023_diffusionwithoutattention} have used bidirectional scanning within the SSM framework. This approach has been expanded to include additional scanning directions~\cite{liu2024swinmamba,liu2024vmamba,yang2024vivim} to account for the characteristics of 2D image data. These approaches unfold image patches along four directions, resulting in four distinct sequences. Each of these sequences is subsequently processed together through every SSM. However, since each direction may have different SSM parameters ($\mathbf{A}$, $\mathbf{B}$, $\mathbf{C}$, and $\mathbf{D}$), scaling up the number of directions could potentially lead to memory issues. 
In this work, we investigate the potential for amortizing the complexity of the Mamba into each layer of the network.

Our approach centers around the concept of token rearrangement before feeding them into the Forward Scan block. For a given input feature $\mathbf{z}_i$ from layer $i$, the output feature $\mathbf{z}_{i+1}$ of the Forward Scan block after the rearrangement can be expressed as:
\begin{align}
    \bz_{\Omega_{i}} &= \texttt{arrange}(\bz_i, \Omega_{i}), \\
    \Bar{\bz}_{\Omega_{i}} &= \texttt{scan}(\bz_{\Omega_{i}}) \label{eq:s6},\\
    \bz_{i+1} &= \texttt{arrange}(\Bar{\bz}_{\Omega_{i}},\Bar{\Omega}_{i}),
\end{align}
$\Omega_i$ represents the 1D permutation  of layer $i$, which rearranges the order of the patch tokens by $\Omega_i$, and $\Omega_i$ and $\overline{\Omega}_i$ represent the reverse operation. This ensures that both $\mathbf{z}_i$ and $\mathbf{z}_{i+1}$ maintain the sample order of the original image tokens.

\begin{figure*}
    \centering
    \begin{subfigure}[h]{0.2\linewidth}
        \centering
        \includegraphics[scale=0.45]{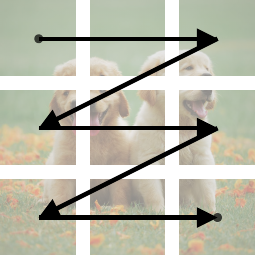}
        \caption{sweep-scan}
    \end{subfigure}
    \begin{subfigure}[h]{0.2\linewidth}
        \centering
        \includegraphics[scale=0.45]{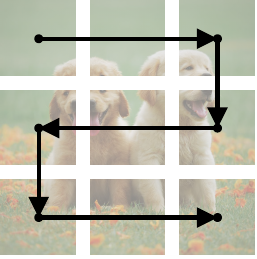}
        \caption{zigzag-scan}
    \end{subfigure}
    \begin{subfigure}[h]{0.5\linewidth}
        \centering
        \includegraphics[scale=0.45]{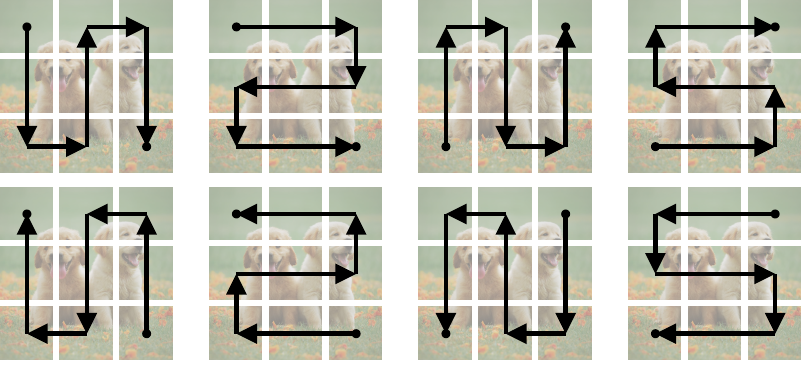}
        \caption{zigzag-scan with 8 schemes}
    \end{subfigure}
    \caption{\textbf{The 2D Image Scan.} Our mamba scan design is based on the sweep-scan scheme shown in subfigure (a). From this, we developed a zigzag-scan scheme displayed in subfigure (b) to enhance the continuity of the patches, thereby maximizing the potential of the Mamba block. Since there are several possible arrangements for these continuous scans, we have listed the eight most common zigzag-scans in subfigure (c).
    }
    \label{fig:img_scan}
\end{figure*}

Now we explore the design of the $\Omega_i$ operation, considering additional inductive biases from 2D images. We propose one key properties: \emph{Spatial Continuity}. Regarding Spatial Continuity,  current innovations of Mamba in images~\cite{visionmamba,liu2024vmamba,liu2024swinmamba} often squeeze 2D patch tokens directly following the computer hierarchy, such as row-and-column-major order. However, this approach may not be optimal for incorporating the inductive bias with neighboring tokens, as illustrated in Figure~\ref{fig:img_scan}. To address this, we introduce a novel scanning scheme designed to maintain spatial continuity during the scan process. 
Additionally, we consider space-filling, which entails that for a patch of size $N \times N$, the length of the 1D continuous scanning scheme should be $N^2$. This helps to efficiently incorporate tokens to maximize the potential of long sequence modeling within the Mamba block.

\noindent\textbf{Heterogeneous Layerwise Scan.} To achieve the aforementioned property, we heuristically design eight possible space-filling continuous schemes\footnote{We also experimented with more complex continuous space-filling paths, such as the Hilbert space-filling curve~\cite{mckenna2019hilbert}. However, empirical findings indicate that this approach may lead to deteriorated results. For further detailed comparisons, please refer to the Appendix.}, denoted as $\bS_j$ (where $j \in [0,7]$), as illustrated in Figure~\ref{fig:img_scan}. While there may be other conceivable schemes, for simplicity, we limit our usage to these eight. Consequently, the scheme for each layer can be represented as $\Omega_i = \bS_{\{i \% 8\}}$, where $\%$ denotes the modulo operator.

\noindent\textbf{Deploying text-condition on Zigzag Mamba.} While Mamba offers the advantage of efficient long sequence modeling, it does so at the expense of the attention mechanism. As a result, there has been limited exploration into incorporating text-conditioning in Mamba-based diffusion models. To address this gap, we propose a straightforward cross-attention block with skip layers built upon the Mamba block, as illustrated in Figure~\ref{fig:block_zigzagmamba}. This design not only enables long sequence modeling but also facilitates multi-token conditioning, such as text-conditioning. Furthermore, it has the potential to provide interpretability~\cite{chefer2021transformer,tang2022daam,p2p}, as cross-attention has been utilized in diffusion models.

\begin{figure*}
    \centering
\includegraphics[width=0.75\textwidth]{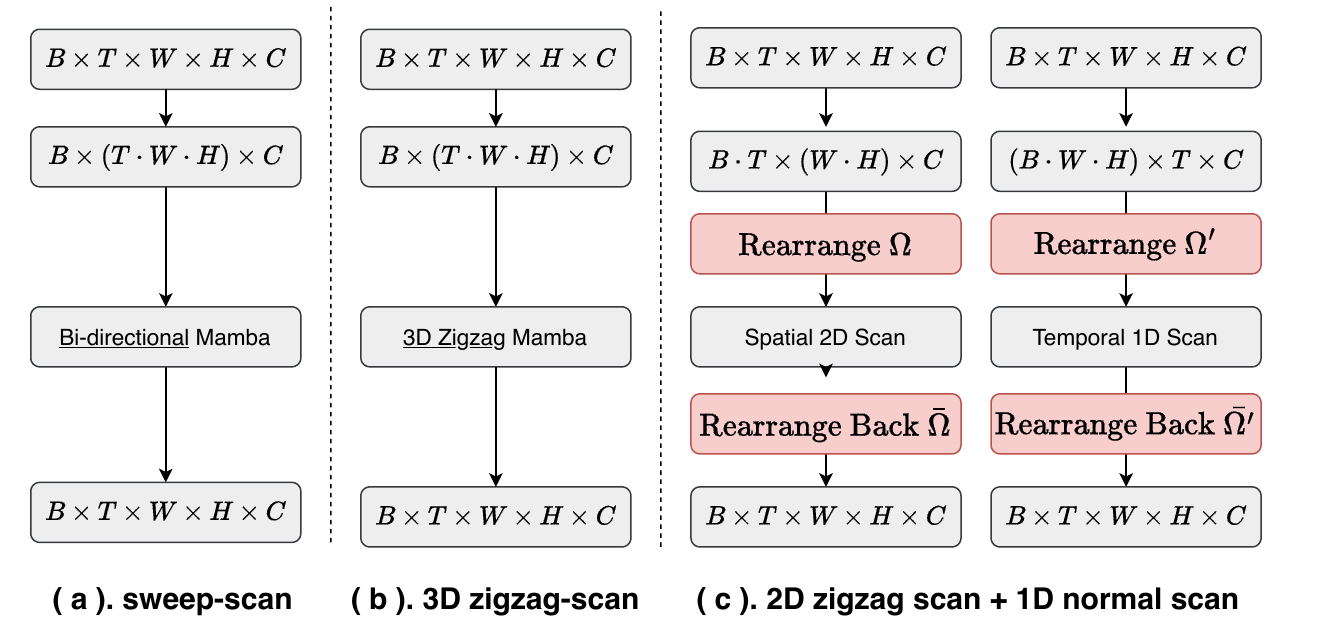}
    \caption{\textbf{The 3D Video Scan.}  \textbf{(a)} We illustrate the bidirectional Mamba with the sweep scan, where the spatial and temporal information is treated as a set of tokens with a computer-hierarchy order. \textbf{(b)} For the 3D zigzag-scan, we aim to maximize the potential of Mamba by employing a spatial continuous scan scheme and adopting the optimal zigzag scan solution, as depicted in Figure~\ref{fig:img_scan}. \textbf{(c)} We further separate the reasoning between spatial and temporal information, resulting in a factorized combination of 2D spatial scan ($\Omega$) plus a 1D temporal scan ($\Omega^{'}$) scheme.
    }
    \label{fig:video_scan}
\end{figure*}

\noindent\textbf{Generalizing to 3D videos by factorizing spatial and temporal information.} In previous sections, our focus has been on the spatial 2D Mamba, where we designed several spatially continuous, space-filling 2D scanning schemes. In this section, we aim to leverage this experience to aid in designing corresponding mechanisms for 3D video processing. We commence our design process by extrapolating from the conventional directional Mamba, as depicted in Figure~\ref{fig:video_scan}. Given a video feature input $\bz\in\mathbb{R}^{B\times T \times C \times W \times H}$, we propose three variants of the Video Mamba Block for facilitating 3D video generation.

(a) Sweep-scan: In this approach, we directly flatten the 3D feature $\mathbf{z}$ without considering spatial or temporal continuity. It's worth noting that the flattening process follows the computer hierarchy order, meaning that no continuity is preserved in the flattened representation.

(b) 3D Zigzag: Compared with the formulation of the 2D zigzag in previous subsections, we follow the similar design to generalize it to 3D Zigzag to keep the continuity in 2D and 3D simultaneously. Potentially, the scheme has much more complexity. We heuristically list 8 schemes as well. However, we empirically find that this scheme will lead to suboptimal optimization.

(c) Factorized 3D Zigzag = 2D Zigzag + 1D Sweep: To address the suboptimal optimization issue, we propose to factorize the spatial and temporal correlations as separate Mamba blocks. The order of their application can be adjusted as desired, for example, "sstt" or "ststst", where "s" represents the spatial-zigzag Mamba and "t" represents the temporal-zigzag Mamba.
For a 1D temporal sweep, we simply opt for forward and backward scanning, since there is only one dimension on the time axis.

\noindent\textbf{Computation Analysis.} %
For a visual sequence $\mathbf{T} \in \mathbb{R}^{1 \times \text{M} \times \text{D}}$, the computation complexity of global self-attention and $k$-direction mamba and our zigzag mamba are as follows:

\begin{align}
\label{eq:self-attn}
&\zeta (\text{self-attention}) = 4\text{M}\text{D}^2 + 2\text{M}^2\text{D}, \\
&\zeta (\text{k-mamba}) = k \times [3\text{M}(2\text{D})\text{N} + \text{M}(2\text{D})\text{N}^2], \\
&\zeta (\text{zigzag}) = 3\text{M}(2\text{D})\text{N} + \text{M}(2\text{D})\text{N}^2,
\end{align}

\noindent where self-attention exhibits quadratic complexity with respect to sequence length $\text{M}$, while Mamba exhibits linear complexity ($\text{N}$ is a fixed parameter, set to 16 by default). Here, $k$ represents the number of scan directions in a single Mamba block. Therefore, $k$-mamba and zigzag share linear complexity with respect to self-attention. Moreover, our zigzag method can eliminate the $k$ series, further reducing the overall complexity.

Upon completing the design of the Zigzag Mamba network for improved visual inductive-bias integration, we proceed to combine it with a new diffusion framework, as illustrated below.

\subsection{Diffusion Framework: Stochastic Interpolant}

\paragraph{Sampling based on vector  $\bv$ and score $\bs$.} Following~\cite{albergo2023stochastic,tong2023simulation}, the time-dependent probability distribution $p_t(\bx)$ of $\bx_t$ also coincides with the distribution of the reverse-time SDE~\cite{anderson1979reverse-time}: 
\begin{equation}
    \label{eq:sde}
    d\bX_t = \bv(\bX_t,t) dt + \frac12w_t \bs(\bX_t,t) dt + \sqrt{w_t} d\bar\bW_t,
\end{equation}
where $\bar\bW_t$ is a reverse-time Wiener process, $w_t>0$ is an arbitrary time-dependent diffusion coefficient,  $\bs(\bx,t)=\nabla \log p_t(\bx)$ is the score, and $\bv(\bx,t)$ is given by the conditional expectation
\begin{equation}
    \label{eq:velocity}
    \begin{aligned}
        \bv(\bx,t) &= \E [ \dot \bx_t | \bx_t = \bx],\\
        &= \dot \alpha_t \E [ \bx_* | \bx_t = \bx]+\dot \sigma_t \E [ \eps | \bx_t = \bx],
    \end{aligned}
\end{equation}
where $\alpha_t$ is a decreasing function of $t$, and $\sigma_t$ is an increasing function of $t$. Here, $\dot{\alpha}_t$ and $\dot{\sigma}_t$ denote the time derivatives of $\alpha_t$ and $\sigma_t$, respectively.

As long as we can estimate the velocity $\bv(\bx, t)$ and/or score $\bs(\bx, t)$ fields, we can utilize it for the sampling process either by  probability flow ODE~\cite{song2021scorebased_sde} or the reverse-time SDE~\eqref{eq:sde}.
Solving the reverse SDE~\eqref{eq:sde} backwards in time from $\bX_{T} = \eps \sim \mathcal{N}(0, \mathbf{I})$ enables generating samples from the approximated data distribution $p_0(\mathbf{x}) \sim p(\bx)$.
During sampling, we can perform direct sampling from either ODE or SDEs to balance between sampling speed and fidelity. If we choose to conduct ODE sampling, we can achieve this simply by setting the noise term $\bs$ to zero. %

In~\cite{albergo2023stochastic}, it shows that one of the two quantities $\bs_\theta(\bx,t)$ and $\bv_\theta(\bx,t)$ needs to be estimated in practice. 
This follows directly from the constraint 
\begin{equation}
\label{eq:c}
\begin{aligned}
    \bx & = \E[\bx_t | \bx_t = \bx],\\
    & = \alpha_t \E[\bx_*|\bx_t=\bx] + \sigma_t \E[\eps|\bx_t=\bx],
\end{aligned}
\end{equation}
which can be used to re-express the score $\bs(\bx, t)$ in terms of the velocity $\bv(\bx,t)$ as
\begin{align}
       \label{eq:v-eps-equivalence}
       \bs(\bx, t) &= \sigma_t^{-1} \frac{\alpha_t \bv(\bx,t) - \dot \alpha_t \bx}{\dot \alpha_t \sigma_t - \alpha_t \dot \sigma_t}.
\end{align}  %
Thus, $\bv(\bx, t)$ and $\bs(\bx, t)$ can be mutually conversed. We illustrate how to compute them in the following.

\label{sec:score:vel}
\noindent\textbf{Estimating the score $\bs$ and the velocity $\bv$.} 
It has been shown in score-based diffusion models~\cite{song2021scorebased_sde} that the score can be estimated parametrically as $\bs_\theta(\bx,t)$ using the loss
\begin{equation}
\label{eq:score:loss}
    \LL_{\mathrm{s}}(\theta) = \int_0^T \E[\Vert \sigma_t \bs_\theta(\bx_t, t) + \eps \Vert^2]\rd t.
\end{equation}
Similarly, the velocity $\bv(\bx,t)$ can be estimated parametrically as $\bv_\theta(\bx,t)$ via the loss
\begin{align}
    \label{eq:velocity-eq-obj}
    \LL_{\mathrm{v}}(\theta) 
    &= \int_0^T \E[\Vert \bv_\theta(\bx_t, t) - \dot\alpha_t \bx_* - \dot\sigma_t \eps\Vert^2] \rd t,
\end{align}
where $\theta$ represents the Zigzag Mamba network that we described in the previous section, we adopt the linear path for training, due to its simplicity and relatively straight trajectory:
\begin{equation}
   \label{eq:alpha:sig}
   \begin{aligned}
 \qquad &\alpha_t = 1-t, \qquad && \sigma_t = t.\\
   \end{aligned}
\end{equation}\textbf{}

We note that any time-dependent weight can be included under the integrals in both~\eqref{eq:score:loss} and~\eqref{eq:velocity-eq-obj}.
These weight factors play a crucial role in score-based models when $T$ becomes large~\cite{kingma2023understanding,kingma2021variational}. Thus, they provide a general form that considers both the time-dependent weight and the stochasticity.

\section{Experiment}

\subsection{Dataset and Training Detail}

\textbf{Image Dataset.} To explore the scalability in high resolution, we conduct experiments on the  FacesHQ $1024\times1024$. The general dataset that we use for training and ablations is FacesHQ, a compilation of CelebA-HQ~\cite{mmcelebahq} and FFHQ~\cite{karras2019_stylegan}, as employed in previous work such as~\cite{esser2021taming,schusterbauer2023boosting}.

\begin{table}
\centering
 \tablestyle{8pt}{1.1}
 \caption{\textbf{Ablation of Scanning Scheme Number}. We evaluate various zigzag scanning schemes. Starting from a simple ``Sweep'' baseline, we consistently observe improvements as more schemes are implemented.}
\begin{tabular}{c|c|c|c||c|c|c}
\toprule
\multicolumn{4}{c||}{\textbf{MultiModal-CelebA-256}} & \multicolumn{3}{c}{ \textbf{MultiModal-CelebA-512}} \\
\specialrule{.1em}{.05em}{.05em} %
  & {FID$^\text{5k}$} $\downarrow$ & {FDD$^\text{5k}$} $\downarrow$ & {KID$^\text{5k}$} $\downarrow$ &{FID$^\text{5k}$} $\downarrow$ & {FDD$^\text{5k}$} $\downarrow$ & {KID$^\text{5k}$} $\downarrow$ \\
\hline
Sweep& 158.1 & 75.9 & 0.169 & {162.3}  &{103.2} &{0.203}  \\
Zigzag-1& 65.7 & 47.8 & 0.051& {121.0}  &  {78.0} &{0.113}  \\
Zigzag-2& 54.7 & 45.5 &0.041 &96.0 &59.5 &0.079 \\
Zigzag-8& 45.5 & 26.4 &0.011 & 34.9  & 29.5 &0.023 \\
\hline
\end{tabular}
    \label{tab:ablation}
\end{table}

\noindent\textbf{Video Dataset.}  UCF101 dataset consists of 13,320 video clips, which are classified into 101 categories. The total length of these video clips is over 27 hours. All these videos are collected from YouTube and have a fixed frame rate of 25 FPS with the resolution of $320\times240$. We randomly sample continuous 16 frames and resize the frames to $256\times 256$.

\noindent\textbf{Training Details.} We uniformly use AdamW~\cite{loshchilov2017decoupled_adamw} optimizer with $1e-4$ learning rate. For extracting latent features, we employ off-the-shelf VAE encoders. To mitigate computational costs, we adopted a mixed-precision training approach.  Additionally, we applied gradient clipping with a threshold of 2.0 and a weight decay of 0.01 to prevent NaN occurrences during Mamba training. Most of our experiments were conducted on 4 A100 GPUs, with scalability exploration extended to 16 and 32 A100 GPUs. For sampling, we adopt the ODE sampling for speed consideration.  For further details, please refer to the Appendix~\ref{supp:details}.

\subsection{Ablation Study}

 \begin{table}
     \centering
     \caption{\textbf{Ablation about Position Embedding (PE)}  on \underline{unconditional} CelebA dataset ($256^2$). To better abate PE and eliminate the conditional signal's influence, we use an unconditional dataset.}
     
     \begin{tabular}{c|ccc}
     \toprule
     \textbf{FID/FDD} $\downarrow$ & \textbf{No PE} & \textbf{Cosine PE} & \textbf{Learnable PE} \\
     \hline 
         VisionMamba~\cite{visionmamba} & 21.33/{21.00}  &  18.47/{19.90} & 16.38/{18.20}  \\
         \text{ZigMa} & \text{14.27/{18.00}} & \text{14.04/{17.91}} & \text{13.32/17.40} \\
         \bottomrule
     \end{tabular}
     \label{tab:pe}
 \end{table}

\begin{figure}
    \centering
    \begin{subfigure}{0.48\textwidth}
        \includegraphics[width=0.9\textwidth]{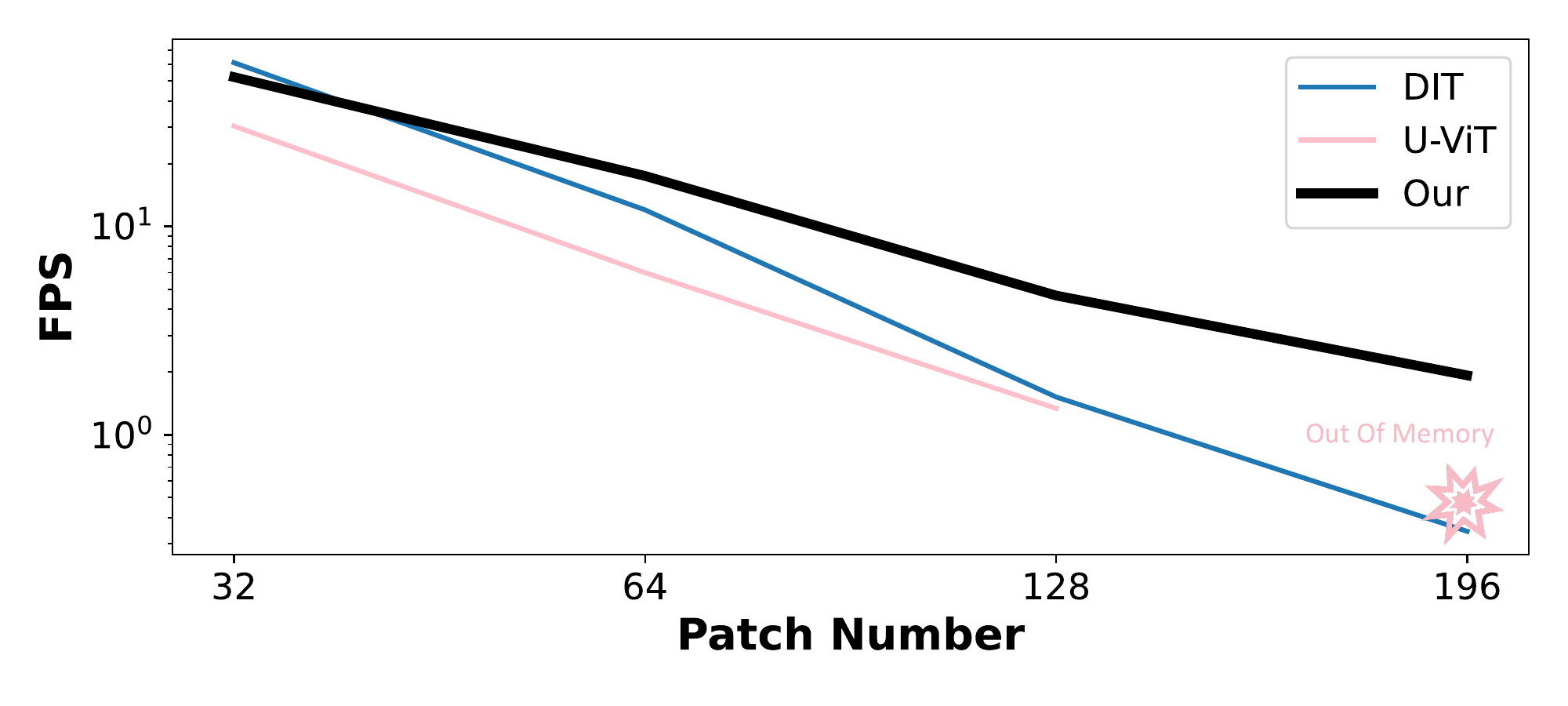}
       \caption{FPS \textit{v.s.} Patch Number.}
    \end{subfigure}
    \hfill
    \begin{subfigure}{0.48\textwidth}
        \includegraphics[width=0.9\textwidth]{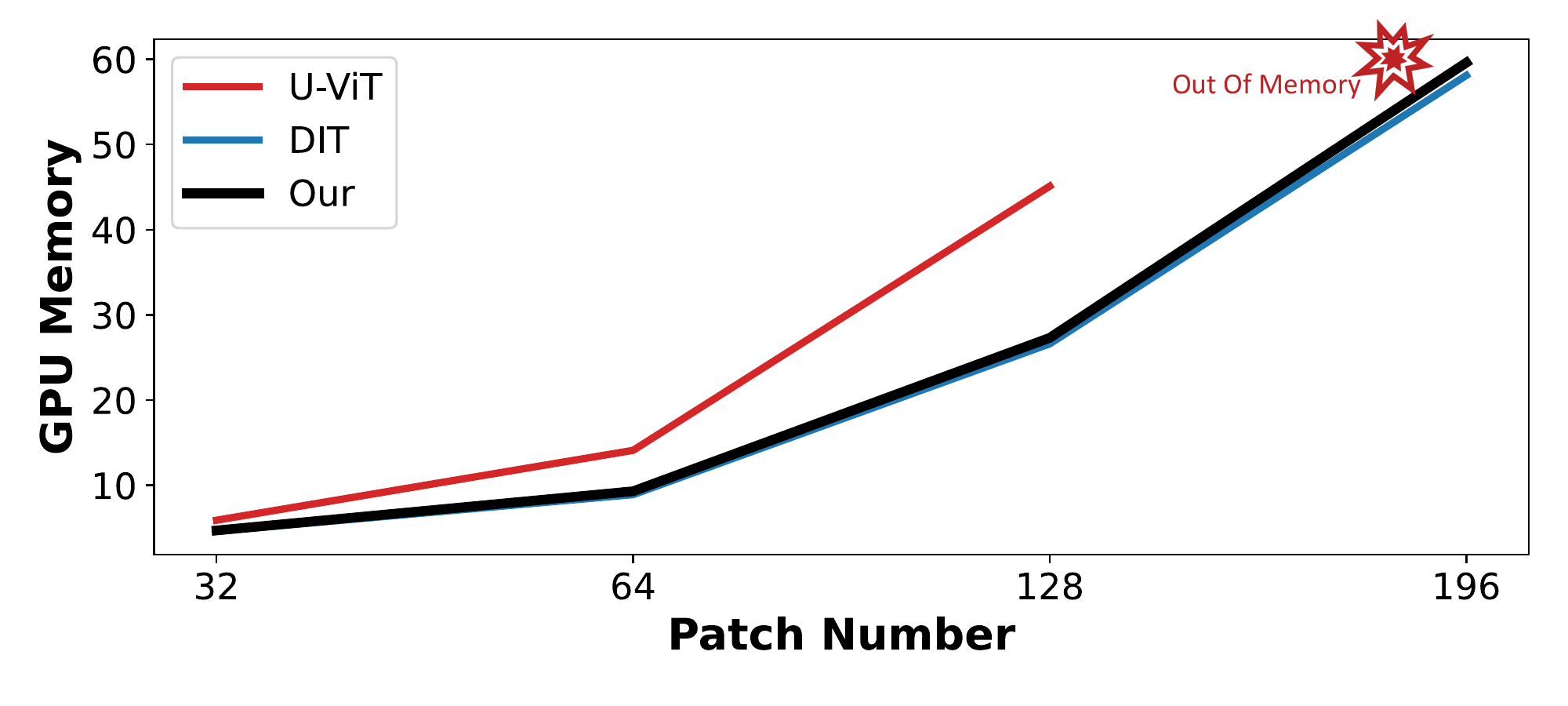}
       \caption{GPU Memory \textit{v.s.} Patch Number.}
    \end{subfigure}
    \vskip\baselineskip
    \begin{subfigure}{0.48\textwidth}
    \includegraphics[width=\textwidth]{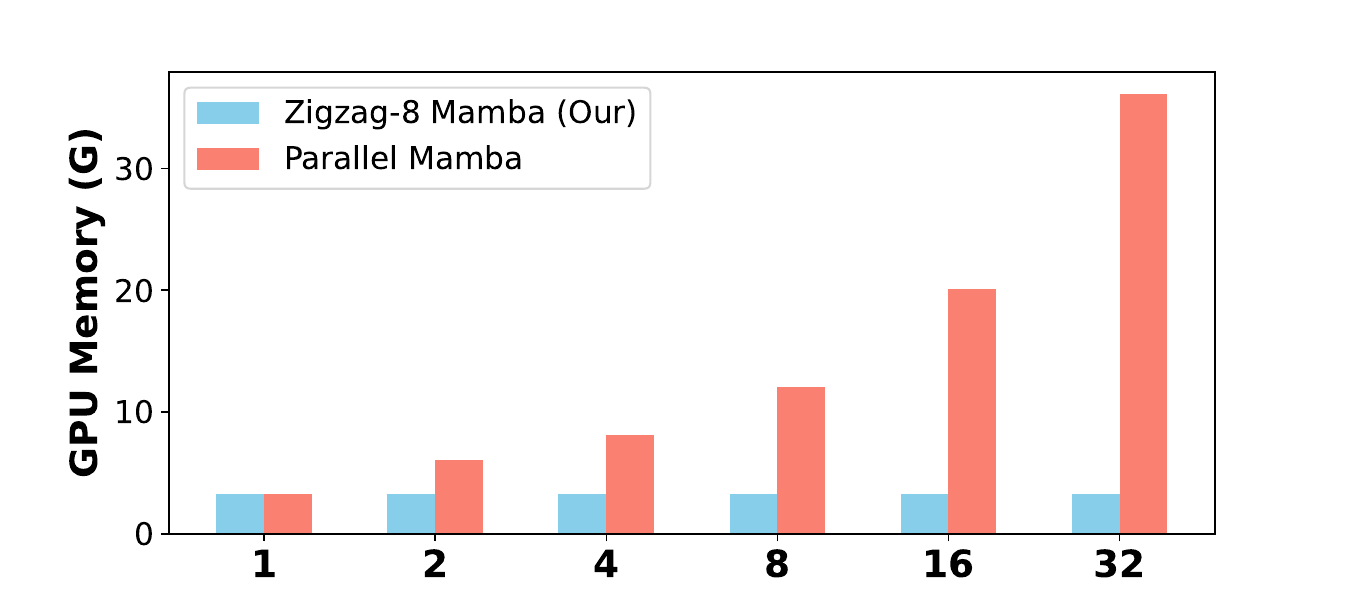}
       \caption{Order Receptive Field \textit{v.s.} GPU Memory.}
    \end{subfigure}
    \hfill
    \begin{subfigure}{0.48\textwidth}
        \includegraphics[trim=0 0 0 30pt, clip, ,width=\textwidth]{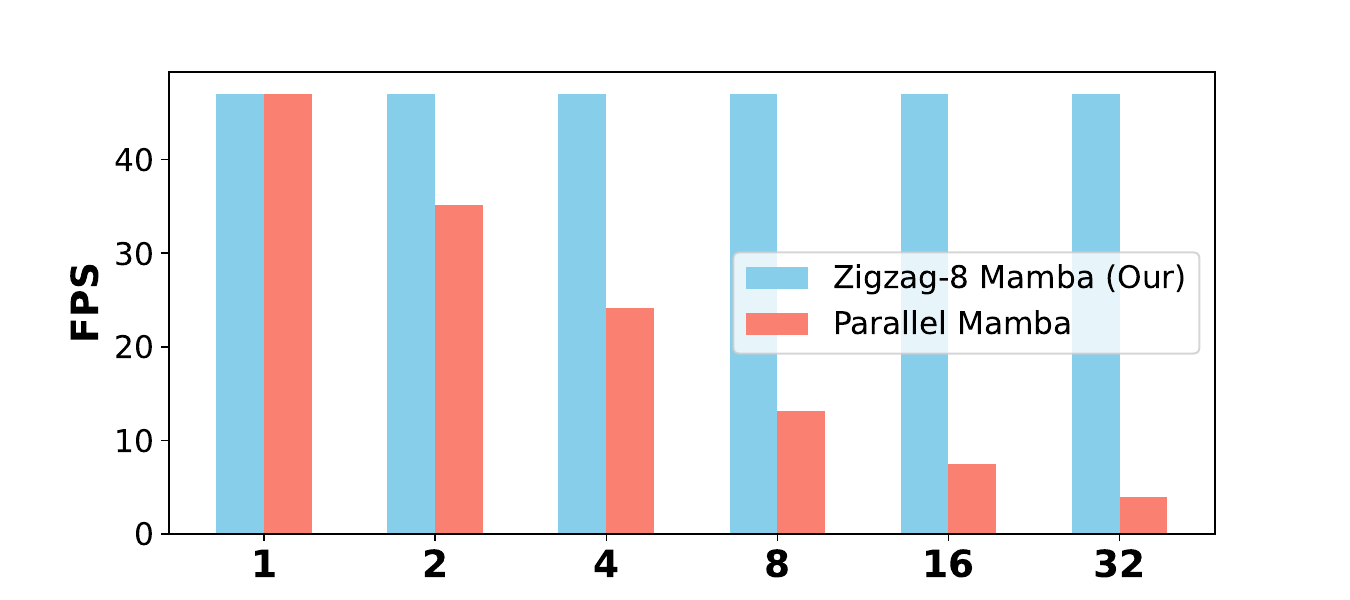}
       \caption{Order Receptive Field \textit{v.s.} FPS.}
    \end{subfigure}

    \caption{(a, b).GPU Memory usage and FPS between our method and transformer-based methods(U-VIT~\cite{uvit} and DiT~\cite{dit_peebles2022scalable}). (c). Order Receptive Field and GPU memory (d). Order Receptive Field and  FPS. Order Receptive Field denotes how many scan paths we consider in our network design.
    }
     \label{fig:main_ablation}
\end{figure}

\noindent\textbf{Scan Scheme Ablation.} We provide several important findings based on our ablation studies on MultiModal-CelebA dataset in various resolutions in Table~\ref{tab:ablation}. 
Firstly, switching the scanning scheme from sweep to zigzag led to some gains. Secondly, as we increased the zigzag scheme from 1 to 8, we saw consistent gains. This indicates that alternating the scanning scheme in various blocks can be beneficial. Finally, the relative gain between Zigzag-1 and Zigzag-8 is more prominent at higher resolutions ($512 \times 512$, or longer  sequence token number) compared to lower resolutions ($256 \times 256$, or shorter sequence token number), this shows the great potential and more efficient inductive-bias incorporation in longer sequence number.

\noindent\textbf{Ablation about Position Embedding.}
As shown in~\Cref{tab:pe}, the learnable embedding performs better than the Sinusoidal embedding, which in turn performs better than no position embedding. In various cases, our zigzag method surpasses the baselines. Notably, our performance remains almost unchanged whether we use the Sinusoidal position embedding or no position embedding. This suggests that our method can better incorporate spatial inductive-bias compared to our baseline. Finally, using the learnable position embedding provides further, albeit marginal, gains suggesting that better position embedding exists even within our zigzag scan scheme. We find that
~\cite{park2024videomamba} shares the same conclusion as us in video-related tasks.

\noindent\textbf{Ablation study about the Network and FPS/GPU-Memory.} In Figure~\ref{fig:main_ablation} (a,b),we analyze the forward speed and GPU memory usage while varying the global patch dimensions from $32\times 32$ to $196 \times 196$.
For the speed analysis, we report Frame Per Second (FPS) instead of FLOPS, as FPS provides a more explicit and appropriate evaluation of speed~\footnote{https://github.com/state-spaces/mamba/issues/110\#issuecomment-1916464012}.
For simplicity, we uniformly apply the zigzag-1 Mamba scan scheme and use batch size=1 and patch size=1 on an A100 GPU with 80GB memory. 
It's worth noting that all methods share nearly identical parameter numbers for fair comparison.
We primarily compare our method with two popular transformer-based Diffusion backbones, U-ViT~\cite{uvit} and DiT~\cite{dit_peebles2022scalable}. It is evident that our method achieves the best FPS and GPU utilization when gradually increasing the patching number. U-ViT demonstrates the worst performance, even exceeds the memory bounds when the patch number is 196. Surprisingly, DiT's GPU utilization is  close to our method, which supports our backbone choice of DiT from a practical perspective.

\begin{table}
    \begin{minipage}{.48\linewidth}
      \caption{\textbf{Main result on FacesHQ-1024 dataset with 4{,}094 tokens in latent space and bs=512.} Our method can outperform the baseline and can achieve even better results when the training scale is increased. }
      \label{tab:faceshq_1024}
      \centering
        \begin{tabular}{@{}l|lll@{}}
    \toprule
        \textbf{Method} & \multicolumn{1}{c}{\textbf{FID}$^\text{5k}$}$\downarrow$   & {\textbf{FDD}$^\text{5k}$} $\downarrow$  \\
        \shline
         VisionMamba~\cite{visionmamba} & 51.1& 66.3 \\
        ZigMa & {37.8}  &  50.5  \\
         \hline  
        ZigMa bs $\times$ 2 & 26.6  &  31.2  \\
        \bottomrule
    \end{tabular}
    \end{minipage}%
    \hspace{\fill}
    \begin{minipage}{.45\linewidth}
      \centering
        \caption{\textbf{Main Results on MS-COCO dataset with bs=256.} Our method consistently outperforms the baseline. ZigMa with 8 scans performs much better compared with the baseline.
        }
     \label{tab:mscoco}
    \begin{tabular}{@{}l|l@{}}
    \toprule
        \textbf{Method} & \multicolumn{1}{c}{\textbf{FID}$^\text{5k}$} $\downarrow$ \\
        \shline
         Sweep & 195.1\\
         Zigzag-1 & 73.1 \\
          VisionMamba~\cite{visionmamba} & 60.2   \\
          \hline 
          Zigzag-8& 41.8   \\
       \bottomrule
    \end{tabular}
    \end{minipage} 
    \begin{minipage}{.48\linewidth}
      \caption{\textbf{Transformer-based methods comparison on unconditional CelebA256.}
      }
      \label{tab:more_flops_memory}
      \centering
       \resizebox{0.95\textwidth}{!}{
        \begin{tabular}{c|ccc}
        \toprule
     \textbf{Method} & \textbf{FID}$\downarrow$ & \textbf{Memory(G)} $\downarrow$ & \textbf{FLOPS(G)} $\downarrow$ \\
        \hline
       U-ViT &{14.50}& {35.10}&12.5   \\
        DiT &{14.64}&{29.20}& 5.5 \\ 
        \hline
          ZigMa &14.27 & 17.80&5.2 \\ 
        \bottomrule
    \end{tabular}
    }
    \end{minipage}%
    \hspace{\fill}
    \begin{minipage}{.43\linewidth}
      \centering
        \caption{\textbf{Video Scan Scheme on UCF101 dataset with bs=32.} 
        }
     \label{tab:ucf101}
     \resizebox{0.96\textwidth}{!}{
    \begin{tabular}{@{}l|cc@{}}
    \toprule
        \textbf{Method} & \multicolumn{1}{c}{\textbf{Frame-FID}$^\text{5k}$ $\downarrow$} & {\textbf{FVD}$^\text{5k}$$\downarrow$}  \\
        \shline
        Bidirection~\cite{visionmamba} & 256.1 &  {320.2} \\
          3D Zigzag &  {238.1} & {282.3}  \\
        Our &  216.1 & {210.2}  \\
        \hline 
         Bidirection~\cite{visionmamba} bs$\times$4 & 146.2 & {201.1}  \\
       ZigMa bs$\times$4& 121.2  & {140.1}  \\
       \bottomrule
    \end{tabular}
    }
    \end{minipage} 
    
\end{table}

\noindent\textbf{Order Receptive Field.} We propose a new concept in Mamba-based structure for multidimensional data. Given that various spatially-continuous zigzag paths may exist in multidimensional data, we introduce the term \emph{Order Receptive Field} which denotes the number of zigzag paths \emph{explicitly} employed in the network design.

\noindent\textbf{Ablation study about the Order Receptive Field and  FPS/GPU-Memory.} As depicted in~\cref{fig:main_ablation} (c,d), Zigzag Mamba consistently maintains its GPU memory consumption and FPS rate, even with a gradually increasing Order Receptive Field. In contrast, our primary baseline, Parallel Mamba, along with variants like Bidirectional Mamba and Vision Mamba~\cite{liu2024vmamba,visionmamba}, experience a consistent decrease in FPS due to increased parameters. Notably, Zigzag Mamba, with an Order Receptive Field of 8, can perform faster without altering parameters.

\noindent\textbf{Comparison with transformer-based methods.} We show the result in~\Cref{tab:more_flops_memory} on unconditional generation task.Our method achieves performance comparable to Transformer-based methods, with significantly less memory consumption and fewer FLOPS.

\subsection{Main Result}

\noindent\textbf{Main Result on 1024$\times$1024 FacesHQ.}  To elaborate on the scalability of our method within the Mamba and Stochastic Interpolant framework, we provide comparisons on a high-resolution dataset (1024$\times$1024 FacesHQ) in Table~\ref{tab:faceshq_1024}. Our primary comparison is against Bidirectional Mamba, a commonly used solution for applying Mamba to 2D image data~\cite{liu2024vmamba,visionmamba}.  With the aim of investigating Mamba's scalability in large resolutions up to 1{,}024, we employ the diffusion model on the latent space of $128 \times 128$ with a patch size of 2, resulting in 4{,}096 tokens. The network is trained on 16 A100 GPUs. Notably, our method demonstrates superior results compared to Bidirectional Mamba. Details regarding loss, FID curves, and visualization can be found in the Appendix. 
While constrained by GPU resource limitations, preventing longer training duration, we anticipate consistent outperformance of Bidirectional Mamba with extended training duration.

\noindent\textbf{COCO dataset.} To further compare the performance of our method, we also evaluate it on the more complex and common dataset MS COCO. We compare with the Bidirection Mamba as the baseline in Table~\ref{tab:mscoco}. It should be noted that all methods share nearly identical parameter numbers for fair comparison.
We trained all methods using 16 A100 GPUs. please check Appendix~\ref{supp:details} for details. As depicted in Table~\ref{tab:mscoco}, our Zigzag-8 method outperforms Bidirectional Mamba as well as Zigzag-1. This suggests that amortizing various scanning schemes can yield significant improvements, attributed to better incorporation of the inductive bias for 2D images in Mamba.

\noindent\textbf{UCF101 dataset.} In Table~\ref{tab:ucf101}, we present our results on the UCF101 dataset, training all methods using 4 A100 GPUs, with further scalability exploration conducted using 16 A100 GPUs. 
We mainly compare our method consistantly with Vision Mamba~\cite{visionmamba}.
For the choice of the 3D Zigzag Mamba, please refer to Appendix~\ref{supp:details}.
For Factorized 3D Zigzag Mamba in video processing, we deploy the \emph{sst} scheme for factorizing spatial and temporal modeling. This scheme prioritizes spatial information complexity over temporal information, hypothesizing that redundancy exists in the temporal domain. Our results consistently demonstrate the superior performance of our method across various scenarios, underscoring the intricacy and effectiveness of our approach.

\section{Conclusion}

In this paper, we present the Zigzag Mamba Diffusion Model, developed within the Stochastic Interpolant framework. Our initial focus is on addressing the critical issue of spatial continuity. We then devise a Zigzag Mamba block with heterogeneous layerwise scan to better utilize the inductive bias in 2D images. Further, we factorize the 3D Mamba into 2D and 1D Zigzag Mamba to facilitate optimization. We empirically design various ablation studies to examine different factors. This approach allows for a more in-depth exploration of the Stochastic Interpolant theory. We hope our endeavor can inspire further exploration in the Mamba network design.

\section*{Acknowledgements}

This project has been supported by the German Federal Ministry for Economic Affairs and Climate Action within the project “NXT GEN AI METHODS – Generative Methoden für Perzeption, Prädiktion und Planung”, the bidt project KLIMA-MEMES, Bayer AG, and the German Research Foundation (DFG) project 421703927. The authors gratefully acknowledge the Gauss Center for Supercomputing for providing compute through the NIC on JUWELS at JSC and the HPC resources supplied by the Erlangen National High Performance Computing Center (NHR@FAU funded by DFG).

\bibliographystyle{splncs04}
\bibliography{mybib}

\begin{thebibliography}{100}
\providecommand{\url}[1]{\texttt{#1}}
\providecommand{\urlprefix}{URL }
\providecommand{\doi}[1]{https://doi.org/#1}

\bibitem{agarwal2023spectral}
Agarwal, N., Suo, D., Chen, X., Hazan, E.: Spectral state space models. arXiv  (2023)

\bibitem{ahamed2024mambatab}
Ahamed, M.A., Cheng, Q.: Mambatab: A simple yet effective approach for handling tabular data. arXiv  (2024)

\bibitem{albergo2023stochastic}
Albergo, M.S., Boffi, N.M., Vanden-Eijnden, E.: Stochastic interpolants: A unifying framework for flows and diffusions. arXiv  (2023)

\bibitem{albergo2022building}
Albergo, M.S., Vanden-Eijnden, E.: Building normalizing flows with stochastic interpolants. arXiv  (2022)

\bibitem{ali2024hidden}
Ali, A., Zimerman, I., Wolf, L.: The hidden attention of mamba models. arXiv  (2024)

\bibitem{anderson1979reverse-time}
Anderson, B.D.: Reverse-time diffusion equation models. Stochastic Processes and their Applications  (1982)

\bibitem{anthony2024blackmamba}
Anthony, Q., Tokpanov, Y., Glorioso, P., Millidge, B.: Blackmamba: Mixture of experts for state-space models. arXiv  (2024)

\bibitem{ao2024burstattention}
Ao, S., Zhao, W., Han, X., Yang, C., Liu, Z., Shi, C., Sun, M., Wang, S., Su, T.: Burstattention: An efficient distributed attention framework for extremely long sequences. arXiv  (2024)

\bibitem{uvit}
Bao, F., Li, C., Cao, Y., Zhu, J.: All are worth words: a vit backbone for score-based diffusion models. CVPR  (2023)

\bibitem{bao2023one}
Bao, F., Nie, S., Xue, K., Li, C., Pu, S., Wang, Y., Yue, G., Cao, Y., Su, H., Zhu, J.: One transformer fits all distributions in multi-modal diffusion at scale. arXiv  (2023)

\bibitem{beck2024xlstm}
Beck, M., Pöppel, K., Spanring, M., Auer, A., Prudnikova, O., Kopp, M., Klambauer, G., Brandstetter, J., Hochreiter, S.: xlstm: Extended long short-term memory (2024)

\bibitem{behrouz2024graphmamba}
Behrouz, A., Hashemi, F.: Graph mamba: Towards learning on graphs with state space models. arXiv  (2024)

\bibitem{beltagy2020longformer}
Beltagy, I., Peters, M.E., Cohan, A.: Longformer: The long-document transformer. arXiv  (2020)

\bibitem{benhamu2022}
Ben-Hamu, H., Cohen, S., Bose, J., Amos, B., Grover, A., Nickel, M., Chen, R.T., Lipman, Y.: Matching normalizing flows and probability paths on manifolds. In: ICML (2022)

\bibitem{brandon2023striped}
Brandon, W., Nrusimha, A., Qian, K., Ankner, Z., Jin, T., Song, Z., Ragan-Kelley, J.: Striped attention: Faster ring attention for causal transformers. arXiv preprint arXiv:2311.09431  (2023)

\bibitem{chefer2021transformer}
Chefer, H., Gur, S., Wolf, L.: Transformer interpretability beyond attention visualization. In: CVPR (2021)

\bibitem{neuralode_chen2018neural}
Chen, R.T., Rubanova, Y., Bettencourt, J., Duvenaud, D.K.: Neural ordinary differential equations. NeurIPS  (2018)

\bibitem{chen2023gentron}
Chen, S., Xu, M., Ren, J., Cong, Y., He, S., Xie, Y., Sinha, A., Luo, P., Xiang, T., Perez-Rua, J.M.: Gentron: Delving deep into diffusion transformers for image and video generation. arXiv  (2023)

\bibitem{child2019generating}
Child, R., Gray, S., Radford, A., Sutskever, I.: Generating long sequences with sparse transformers. arXiv  (2019)

\bibitem{choromanski2020rethinking}
Choromanski, K., Likhosherstov, V., Dohan, D., Song, X., Gane, A., Sarlos, T., Hawkins, P., Davis, J., Mohiuddin, A., Kaiser, L., et~al.: Rethinking attention with performers. arXiv  (2020)

\bibitem{crowson2024scalable_hdit}
Crowson, K., Baumann, S.A., Birch, A., Abraham, T.M., Kaplan, D.Z., Shippole, E.: Scalable high-resolution pixel-space image synthesis with hourglass diffusion transformers. arXiv  (2024)

\bibitem{dao2023flow_lfm}
Dao, Q., Phung, H., Nguyen, B., Tran, A.: Flow matching in latent space. arXiv  (2023)

\bibitem{dao2022flashattention}
Dao, T., Fu, D., Ermon, S., Rudra, A., R{\'e}, C.: Flashattention: Fast and memory-efficient exact attention with io-awareness. NeurIPS  (2022)

\bibitem{dehghani2023scaling}
Dehghani, M., Djolonga, J., Mustafa, B., Padlewski, P., Heek, J., Gilmer, J., Steiner, A.P., Caron, M., Geirhos, R., Alabdulmohsin, I., et~al.: Scaling vision transformers to 22 billion parameters. In: ICML (2023)

\bibitem{dosovitskiy2020image_vit}
Dosovitskiy, A., Beyer, L., Kolesnikov, A., Weissenborn, D., Zhai, X., Unterthiner, T., Dehghani, M., Minderer, M., Heigold, G., Gelly, S., et~al.: An image is worth 16x16 words: Transformers for image recognition at scale. In: ICLR (2021)

\bibitem{esser2021taming}
Esser, P., Rombach, R., Ommer, B.: Taming transformers for high-resolution image synthesis. In: CVPR (2021)

\bibitem{fei2024scalable_dis}
Fei, Z., Fan, M., Yu, C., Huang, J.: Scalable diffusion models with state space backbone. arXiv  (2024)

\bibitem{schusterbauer2023boosting}
Schusterbauer, J.S., Gui, M., Ma, P., Stracke, N., Baumann, S.A., Ommer, B.: Boosting latent diffusion with flow matching. ECCV  (2024)

\bibitem{fu2022hungry_h3}
Fu, D.Y., Dao, T., Saab, K.K., Thomas, A.W., Rudra, A., R{\'e}, C.: Hungry hungry hippos: Towards language modeling with state space models. arXiv  (2022)

\bibitem{fuest2024diffusion}
Fuest, M., Ma, P., Gui, M., Schusterbauer, J.S., Hu, V.T., Ommer, B.: Diffusion models and representation learning: A survey. arXiv preprint arXiv:2407.00783  (2024)

\bibitem{gong2024nnmamba}
Gong, H., Kang, L., Wang, Y., Wan, X., Li, H.: nnmamba: 3d biomedical image segmentation, classification and landmark detection with state space model. arXiv  (2024)

\bibitem{gong2023diffpose}
Gong, J., Foo, L.G., Fan, Z., Ke, Q., Rahmani, H., Liu, J.: Diffpose: Toward more reliable 3d pose estimation. In: CVPR (2023)

\bibitem{gu2023mamba}
Gu, A., Dao, T.: Mamba: Linear-time sequence modeling with selective state spaces. CoLM  (2024)

\bibitem{gu2022parameterization}
Gu, A., Goel, K., Gupta, A., R{\'e}, C.: On the parameterization and initialization of diagonal state space models. NeurIPS  (2022)

\bibitem{gu2021efficiently}
Gu, A., Goel, K., R{\'e}, C.: Efficiently modeling long sequences with structured state spaces (2021)

\bibitem{gu2021combining}
Gu, A., Johnson, I., Goel, K., Saab, K., Dao, T., Rudra, A., R{\'e}, C.: Combining recurrent, convolutional, and continuous-time models with linear state space layers. NeurIPS  (2021)

\bibitem{gui2024depthfm}
Gui, M., Schusterbauer, J.S., Prestel, U., Ma, P., Kotovenko, D., Grebenkova, O., Baumann, S.A., Hu, V.T., Ommer, B.: Depthfm: Fast monocular depth estimation with flow matching. arXiv preprint arXiv:2403.13788  (2024)

\bibitem{guo2024mambair}
Guo, H., Li, J., Dai, T., Ouyang, Z., Ren, X., Xia, S.T.: Mambair: A simple baseline for image restoration with state-space model. arXiv  (2024)

\bibitem{gupta2022diagonal}
Gupta, A., Gu, A., Berant, J.: Diagonal state spaces are as effective as structured state spaces. NeurIPS  (2022)

\bibitem{he2024densemamba}
He, W., Han, K., Tang, Y., Wang, C., Yang, Y., Guo, T., Wang, Y.: Densemamba: State space models with dense hidden connection for efficient large language models. arXiv  (2024)

\bibitem{he2024pan}
He, X., Cao, K., Yan, K., Li, R., Xie, C., Zhang, J., Zhou, M.: Pan-mamba: Effective pan-sharpening with state space model. arXiv  (2024)

\bibitem{p2p}
Hertz, A., Mokady, R., Tenenbaum, J., Aberman, K., Pritch, Y., Cohen-Or, D.: Prompt-to-prompt image editing with cross attention control. arXiv  (2022)

\bibitem{ho2020denoising}
Ho, J., Jain, A., Abbeel, P.: Denoising diffusion probabilistic models. In: NeurIPS (2020)

\bibitem{ho2022video}
Ho, J., Salimans, T., Gritsenko, A., Chan, W., Norouzi, M., Fleet, D.J.: Video diffusion models. In: ARXIV (2022)

\bibitem{sgfm}
Hu, V.T., Chen, Y., Caron, M., Asano, Y.M., Snoek, C.G., Ommer, B.: Guided diffusion from self-supervised diffusion features. In: ARXIV (2023)

\bibitem{flowseq}
Hu, V.T., Wu, D., Asano, Y., Mettes, P., Fernando, B., Ommer, B., Snoek, C.: Flow matching for conditional text generation in a few sampling steps pp. 380--392 (2024)

\bibitem{motionfm}
Hu, V.T., Yin, W., Ma, P., Chen, Y., Fernando, B., Asano, Y.M., Gavves, E., Mettes, P., Ommer, B., Snoek, C.G.: Motion flow matching for human motion synthesis and editing. In: ARXIV (2023)

\bibitem{sgdm}
Hu, V.T., Zhang, D.W., Asano, Y.M., Burghouts, G.J., Snoek, C.G.M.: Self-guided diffusion models. In: CVPR (2023)

\bibitem{hulfm}
Hu, V.T., Zhang, D.W., Mettes, P., Tang, M., Zhao, D., Snoek, C.G.: Latent space editing in transformer-based flow matching. In: ICML 2023 Workshop, New Frontiers in Learning, Control, and Dynamical Systems (2023)

\bibitem{huang2024scalelong}
Huang, Z., Zhou, P., Yan, S., Lin, L.: Scalelong: Towards more stable training of diffusion model via scaling network long skip connection. NeurIPS  (2024)

\bibitem{huang2021shuffle}
Huang, Z., Ben, Y., Luo, G., Cheng, P., Yu, G., Fu, B.: Shuffle transformer: Rethinking spatial shuffle for vision transformer. arXiv preprint arXiv:2106.03650  (2021)

\bibitem{karras2022elucidating}
Karras, T., Aittala, M., Aila, T., Laine, S.: Elucidating the design space of diffusion-based generative models. In: NeurIPS (2022)

\bibitem{karras2019_stylegan}
Karras, T., Laine, S., Aila, T.: A style-based generator architecture for generative adversarial networks. In: CVPR (2019)

\bibitem{kingma2021variational}
Kingma, D., Salimans, T., Poole, B., Ho, J.: Variational diffusion models. In: NeurIPS (2021)

\bibitem{kingma2023understanding}
Kingma, D.P., Gao, R.: Understanding the diffusion objective as a weighted integral of elbos. arXiv  (2023)

\bibitem{kitaev2020reformer}
Kitaev, N., Kaiser, {\L}., Levskaya, A.: Reformer: The efficient transformer. arXiv  (2020)

\bibitem{lee2023minimizing}
Lee, S., Kim, B., Ye, J.C.: Minimizing trajectory curvature of ode-based generative models. ICML  (2023)

\bibitem{li2024videomamba}
Li, K., Li, X., Wang, Y., He, Y., Wang, Y., Wang, L., Qiao, Y.: Videomamba: State space model for efficient video understanding. ECCV  (2024)

\bibitem{li2024mamba_nd}
Li, S., Singh, H., Grover, A.: Mamba-nd: Selective state space modeling for multi-dimensional data. arXiv  (2024)

\bibitem{li2024denoising}
Li, Y., Bornschein, J., Chen, T.: Denoising autoregressive representation learning. arXiv preprint arXiv:2403.05196  (2024)

\bibitem{liang2024pointmamba}
Liang, D., Zhou, X., Wang, X., Zhu, X., Xu, W., Zou, Z., Ye, X., Bai, X.: Pointmamba: A simple state space model for point cloud analysis. arXiv preprint arXiv:2402.10739  (2024)

\bibitem{lin2024mtmamba}
Lin, B., Jiang, W., Chen, P., Zhang, Y., Liu, S., Chen, Y.C.: Mtmamba: Enhancing multi-task dense scene understanding by mamba-based decoders. ECCV  (2024)

\bibitem{coco}
Lin, T.Y., Maire, M., Belongie, S., Hays, J., Perona, P., Ramanan, D., Doll{\'a}r, P., Zitnick, C.L.: Microsoft coco: Common objects in context. In: ECCV (2014)

\bibitem{lipman2022flow}
Lipman, Y., Chen, R.T., Ben-Hamu, H., Nickel, M., Le, M.: Flow matching for generative modeling. ICLR  (2023)

\bibitem{liu2022deep_gdsb}
Liu, G.H., Chen, T., So, O., Theodorou, E.: Deep generalized schr{\"o}dinger bridge. NeurIPS  (2022)

\bibitem{liu2023ring}
Liu, H., Zaharia, M., Abbeel, P.: Ring attention with blockwise transformers for near-infinite context. arXiv  (2023)

\bibitem{liu2024swinmamba}
Liu, J., Yang, H., Zhou, H.Y., Xi, Y., Yu, L., Yu, Y., Liang, Y., Shi, G., Zhang, S., Zheng, H., et~al.: Swin-umamba: Mamba-based unet with imagenet-based pretraining. arXiv  (2024)

\bibitem{liu2022flow}
Liu, X., Gong, C., Liu, Q.: Flow straight and fast: Learning to generate and transfer data with rectified flow. arXiv  (2022)

\bibitem{rectifiedflow_iclr23}
Liu, X., Gong, C., Liu, Q.: Flow straight and fast: Learning to generate and transfer data with rectified flow. ICLR  (2023)

\bibitem{liu2024vmamba}
Liu, Y., Tian, Y., Zhao, Y., Yu, H., Xie, L., Wang, Y., Ye, Q., Liu, Y.: Vmamba: Visual state space model. arXiv  (2024)

\bibitem{swin_transformer}
Liu, Z., Lin, Y., Cao, Y., Hu, H., Wei, Y., Zhang, Z., Lin, S., Guo, B.: Swin transformer: Hierarchical vision transformer using shifted windows. In: ICCV (2021)

\bibitem{loshchilov2017decoupled_adamw}
Loshchilov, I., Hutter, F.: Decoupled weight decay regularization. In: ICLR (2019)

\bibitem{umamba}
Ma, J., Li, F., Wang, B.: U-mamba: Enhancing long-range dependency for biomedical image segmentation. arXiv  (2024)

\bibitem{ma2024sit}
Ma, N., Goldstein, M., Albergo, M.S., Boffi, N.M., Vanden-Eijnden, E., Xie, S.: Sit: Exploring flow and diffusion-based generative models with scalable interpolant transformers. ECCV  (2024)

\bibitem{mckenna2019hilbert}
McKenna, D.M.: Hilbert curves: Outside-in and inside-gone. Mathemaesthetics, Inc  (2019)

\bibitem{hourglass}
Newell, A., Yang, K., Deng, J.: Stacked hourglass networks for human pose estimation. In: ECCV (2016)

\bibitem{nguyen2022s4nd}
Nguyen, E., Goel, K., Gu, A., Downs, G., Shah, P., Dao, T., Baccus, S., R{\'e}, C.: S4nd: Modeling images and videos as multidimensional signals with state spaces. NeurIPS  (2022)

\bibitem{sora}
OpenAI: Sora: Creating video from text (2024), \url{https://openai.com/sora}

\bibitem{park2024videomamba}
Park, J., Kim, H.S., Ko, K., Kim, M., Kim, C.: Videomamba: Spatio-temporal selective state space model. ECCV  (2024)

\bibitem{dit_peebles2022scalable}
Peebles, W., Xie, S.: Scalable diffusion models with transformers. arXiv  (2022)

\bibitem{peng2024eagle_rwkv2}
Peng, B., Goldstein, D., Anthony, Q., Albalak, A., Alcaide, E., Biderman, S., Cheah, E., Ferdinan, T., Hou, H., Kazienko, P., et~al.: Eagle and finch: Rwkv with matrix-valued states and dynamic recurrence. arXiv preprint arXiv:2404.05892  (2024)

\bibitem{qin2024hgrn2}
Qin, Z., Yang, S., Sun, W., Shen, X., Li, D., Sun, W., Zhong, Y.: Hgrn2: Gated linear rnns with state expansion. arXiv preprint arXiv:2404.07904  (2024)

\bibitem{radford2021learning_clip}
Radford, A., Kim, J.W., Hallacy, C., Ramesh, A., Goh, G., Agarwal, S., Sastry, G., Askell, A., Mishkin, P., Clark, J., et~al.: Learning transferable visual models from natural language supervision. In: ICML (2021)

\bibitem{rombach2022high_latentdiffusion_ldm}
Rombach, R., Blattmann, A., Lorenz, D., Esser, P., Ommer, B.: High-resolution image synthesis with latent diffusion models. In: CVPR (2022)

\bibitem{unet}
Ronneberger, O., Fischer, P., Brox, T.: U-net: Convolutional networks for biomedical image segmentation. In: MICCAI (2015)

\bibitem{ruan2024vm}
Ruan, J., Xiang, S.: Vm-unet: Vision mamba unet for medical image segmentation. arXiv  (2024)

\bibitem{skorokhodov2021aligning_landscapehq}
Skorokhodov, I., Sotnikov, G., Elhoseiny, M.: Aligning latent and image spaces to connect the unconnectable. In: ICCV (2021)

\bibitem{smith2022simplified_s4}
Smith, J.T., Warrington, A., Linderman, S.W.: Simplified state space layers for sequence modeling. arXiv  (2022)

\bibitem{sohl2015deep}
Sohl-Dickstein, J., Weiss, E., Maheswaranathan, N., Ganguli, S.: Deep unsupervised learning using nonequilibrium thermodynamics. In: ICML (2015)

\bibitem{song2019generative}
Song, Y., Ermon, S.: Generative modeling by estimating gradients of the data distribution. arXiv  (2019)

\bibitem{song2021scorebased_sde}
Song, Y., Sohl-Dickstein, J., Kingma, D.P., Kumar, A., Ermon, S., Poole, B.: Score-based generative modeling through stochastic differential equations. In: ICLR (2021)

\bibitem{stein2024exposing}
Stein, G., Cresswell, J., Hosseinzadeh, R., Sui, Y., Ross, B., Villecroze, V., Liu, Z., Caterini, A.L., Taylor, E., Loaiza-Ganem, G.: Exposing flaws of generative model evaluation metrics and their unfair treatment of diffusion models. NeurIPS  (2023)

\bibitem{sun2021sparse}
Sun, Z., Yang, Y., Yoo, S.: Sparse attention with learning to hash. In: ICLR (2021)

\bibitem{tang2022daam}
Tang, R., Liu, L., Pandey, A., Jiang, Z., Yang, G., Kumar, K., Stenetorp, P., Lin, J., Ture, F.: What the daam: Interpreting stable diffusion using cross attention. arXiv  (2022)

\bibitem{tikochinski2024incremental}
Tikochinski, R., Goldstein, A., Meiri, Y., Hasson, U., Reichart, R.: An incremental large language model for long text processing in the brain  (2024)

\bibitem{tong2023simulation}
Tong, A., Malkin, N., Fatras, K., Atanackovic, L., Zhang, Y., Huguet, G., Wolf, G., Bengio, Y.: Simulation-free schr$\backslash$" odinger bridges via score and flow matching. arXiv  (2023)

\bibitem{unterthiner2019fvd}
Unterthiner, T., van Steenkiste, S., Kurach, K., Marinier, R., Michalski, M., Gelly, S.: Fvd: A new metric for video generation. ICLR Workshop  (2019)

\bibitem{transformer}
Vaswani, A., Shazeer, N., Parmar, N., Uszkoreit, J., Jones, L., Gomez, A.N., Kaiser, {\L}., Polosukhin, I.: Attention is all you need. In: NeurIPS (2017)

\bibitem{wang2024graph}
Wang, C., Tsepa, O., Ma, J., Wang, B.: Graph-mamba: Towards long-range graph sequence modeling with selective state spaces. arXiv  (2024)

\bibitem{wang2024mambabyte}
Wang, J., Gangavarapu, T., Yan, J.N., Rush, A.M.: Mambabyte: Token-free selective state space model. arXiv  (2024)

\bibitem{bigs_wang2022pretraining}
Wang, J., Yan, J.N., Gu, A., Rush, A.M.: Pretraining without attention. arXiv  (2022)

\bibitem{wang2023stablessm}
Wang, S., Li, Q.: Stablessm: Alleviating the curse of memory in state-space models through stable reparameterization. arXiv  (2023)

\bibitem{wang2024state}
Wang, S., Xue, B.: State-space models with layer-wise nonlinearity are universal approximators with exponential decaying memory. NeurIPS  (2024)

\bibitem{1024words}
Wang, W., Ma, S., Xu, H., Usuyama, N., Ding, J., Poon, H., Wei, F.: When an image is worth 1,024 x 1,024 words: A case study in computational pathology. arXiv  (2023)

\bibitem{wang2024state_survey}
Wang, X., Wang, S., Ding, Y., Li, Y., Wu, W., Rong, Y., Kong, W., Huang, J., Li, S., Yang, H., Wang, Z., Jiang, B., Li, C., Wang, Y., Tian, Y., Tang, J.: State space model for new-generation network alternative to transformers: A survey (2024)

\bibitem{wang2024text}
Wang, X., Kang, Z., Mu, Y.: Text-controlled motion mamba: Text-instructed temporal grounding of human motion. arXiv preprint arXiv:2404.11375  (2024)

\bibitem{wang2024semi}
Wang, Z., Ma, C.: Semi-mamba-unet: Pixel-level contrastive cross-supervised visual mamba-based unet for semi-supervised medical image segmentation. arXiv  (2024)

\bibitem{wang2024mamba}
Wang, Z., Zheng, J.Q., Zhang, Y., Cui, G., Li, L.: Mamba-unet: Unet-like pure visual mamba for medical image segmentation. arXiv  (2024)

\bibitem{wu2023fast}
Wu, L., Wang, D., Gong, C., Liu, X., Xiong, Y., Ranjan, R., Krishnamoorthi, R., Chandra, V., Liu, Q.: Fast point cloud generation with straight flows. In: CVPR (2023)

\bibitem{mmcelebahq}
Xia, W., Yang, Y., Xue, J.H., Wu, B.: Tedigan: Text-guided diverse face image generation and manipulation. In: CVPR (2021)

\bibitem{xing2024segmamba}
Xing, Z., Ye, T., Yang, Y., Liu, G., Zhu, L.: Segmamba: Long-range sequential modeling mamba for 3d medical image segmentation. arXiv  (2024)

\bibitem{yan2023_diffusionwithoutattention}
Yan, J.N., Gu, J., Rush, A.M.: Diffusion models without attention. arXiv  (2023)

\bibitem{yang2023gated_gla}
Yang, S., Wang, B., Shen, Y., Panda, R., Kim, Y.: Gated linear attention transformers with hardware-efficient training. ICML  (2024)

\bibitem{yang2024fla}
Yang, S., Zhang, Y.: Fla: A triton-based library for hardware-efficient implementations of linear attention mechanism (Jan 2024), \url{https://github.com/sustcsonglin/flash-linear-attention}

\bibitem{yang2024vivim}
Yang, Y., Xing, Z., Zhu, L.: Vivim: a video vision mamba for medical video object segmentation. arXiv  (2024)

\bibitem{yu2023robustifying}
Yu, A., Nigmetov, A., Morozov, D., Mahoney, M.W., Erichson, N.B.: Robustifying state-space models for long sequences via approximate diagonalization. arXiv  (2023)

\bibitem{yu2023video}
Yu, S., Sohn, K., Kim, S., Shin, J.: Video probabilistic diffusion models in projected latent space. In: CVPR (2023)

\bibitem{zhang2024point}
Zhang, T., Li, X., Yuan, H., Ji, S., Yan, S.: Point could mamba: Point cloud learning via state space model. arXiv  (2024)

\bibitem{zhang2018shufflenet}
Zhang, X., Zhou, X., Lin, M., Sun, J.: Shufflenet: An extremely efficient convolutional neural network for mobile devices. In: CVPR (2018)

\bibitem{zhang2024motionmamba}
Zhang, Z., Liu, A., Reid, I., Hartley, R., Zhuang, B., Tang, H.: Motion mamba: Efficient and long sequence motion generation with hierarchical and bidirectional selective ssm. ECCV  (2024)

\bibitem{motionmamba}
Zhang, Z., Liu, A., Reid, I., Hartley, R., Zhuang, B., Tang, H.: Motion mamba: Efficient and long sequence motion generation with hierarchical and bidirectional selective ssm. arXiv  (2024)

\bibitem{zheng2024u_mamba_dehazing}
Zheng, Z., Wu, C.: U-shaped vision mamba for single image dehazing. arXiv  (2024)

\bibitem{visionmamba}
Zhu, L., Liao, B., Zhang, Q., Wang, X., Liu, W., Wang, X.: Vision mamba: Efficient visual representation learning with bidirectional state space model. ICML  (2024)

\bibitem{ring_flash_attention}
zhuzilin: Ring flash attention. \url{https://github.com/zhuzilin/ring-flash-attention} (2024)

\end{thebibliography}

\clearpage

\section{Limitations and Future Work}

Our method relies solely on the Mamba Block with a DiT-style layout and conditioning manner. However, a potential limitation of our work is that we cannot exhaustively list all possible spatial continuous zigzag scanning schemes given a specific global patch size. Currently, we set these scanning schemes empirically, which may lead to sub-optimal performance. Additionally, due to GPU resource constraints, we were unable to explore longer training durations, although we anticipate similar conclusions.

For future work, we aim to delve into various applications of the Zigzag Mamba, leveraging its scalability for long-sequence modeling. This exploration may lead to improved utilization of the Mamba framework across different domains and applications.

Ultimately, we anticipate that our scan path will be suitable for other linear attention models such as RWKV~\cite{peng2024eagle_rwkv2}, xLSTM~\cite{beck2024xlstm}, HGRN~\cite{qin2024hgrn2}, GLA~\cite{yang2023gated_gla}, and several others listed at FLA~\cite{yang2024fla}\footnote{\url{https://github.com/sustcsonglin/flash-linear-attention}}.

\section{Impact Statement}

This work aims to enhance the scalability and unlock the potential of the Mamba algorithm within the framework of diffusion models, enabling the generation of large images with high-fidelity. By incorporating our cross-attention mechanism into the Mamba block, our method can also facilitate text-to-image generation. However, like other endeavors aimed at enhancing the capabilities and control of large-scale image synthesis models, our approach carries the risk of enabling the generation of harmful or deceptive content. Therefore, ethical considerations and safeguards must be implemented to mitigate these risks.

\section{Appendix}

\begin{table}
    \tablestyle{8pt}{1.1}
    \caption{\textbf{The ablation about Hilbert and Zigzag scan path under various Order Receptive Field ( ORF )  on unconditional MultiModal-CelebA256.}  }
    \begin{tabular}{@{}c|lll@{}}
        Scan-ORF & \multicolumn{1}{c}{FID$^\text{5k}$} \\
        \shline
         hilbert-2 & 61.67\\
         hilbert-8 & 27.38\\
        \hline
         zigzag-2 & 15.45\\
        zigzag-8 &  13.32  \\
    \end{tabular}
    \vspace{.5em}
    \label{tab:ablate_hilbert}
\vspace{-1.5em}
\end{table}

\begin{table}
    \tablestyle{8pt}{1.1}
    \caption{\textbf{Various methods for text-to-image generation on the MultiModal-CelebA 256 dataset.}}
    \begin{tabular}{@{}l|lll@{}}
        Method & \multicolumn{1}{c}{FID$^\text{5k}$} & {FDD$^\text{5k}$} & {KID$^\text{5k}$} \\
        \shline
        In-Context &  {61.1} & {39.1} & {0.061} \\
        Cross-Attention & 45.5 & 26.4 & 0.011  \\
    \end{tabular}
    \vspace{.5em}
    \label{tab:cross_attention}
\vspace{-1.5em}
\end{table}

\begin{table}
\centering
\caption{\textbf{Details of ZigMa Model Variants.} We follow previous works~\cite{dit_peebles2022scalable,uvit,dosovitskiy2020image_vit} model configurations for the Small (S), Base (B) and Large (L) variants; we also introduce an XLarge (XL) config as our largest model. CA denotes the cross-attention for text-to-image conditioning.}
\scalebox{1.0}{
\begin{tabular}{l c c c}
Model            & Layers $N$ & Hidden size $d$ & \#params \\
\toprule 
ZigMa-S   &   12   &     384  &31.3M       \\
ZigMa-B   &   12   &    768    & 133.8M       \\
ZigMa-L  &    24   &      1024  & 472.5M       \\
ZigMa-XL &    28  &       1152  &1058.7M       \\
\hline
CA-ZigMa-S   &   12   &     384  &59.2M       \\
CA-ZigMa-B   &   12   &    768    & 214.1M       \\
CA-ZigMa-L  &    24   &      1024  & 724.4M       \\
CA-ZigMa-XL &    28  &       1152  &1549.8M       \\
\bottomrule
\end{tabular}
}
\label{tab:models}
\end{table}

\subsection{Visualization}
\label{supp:sec:vis}

\begin{figure*}
    \centering
\includegraphics[width=0.99\textwidth]{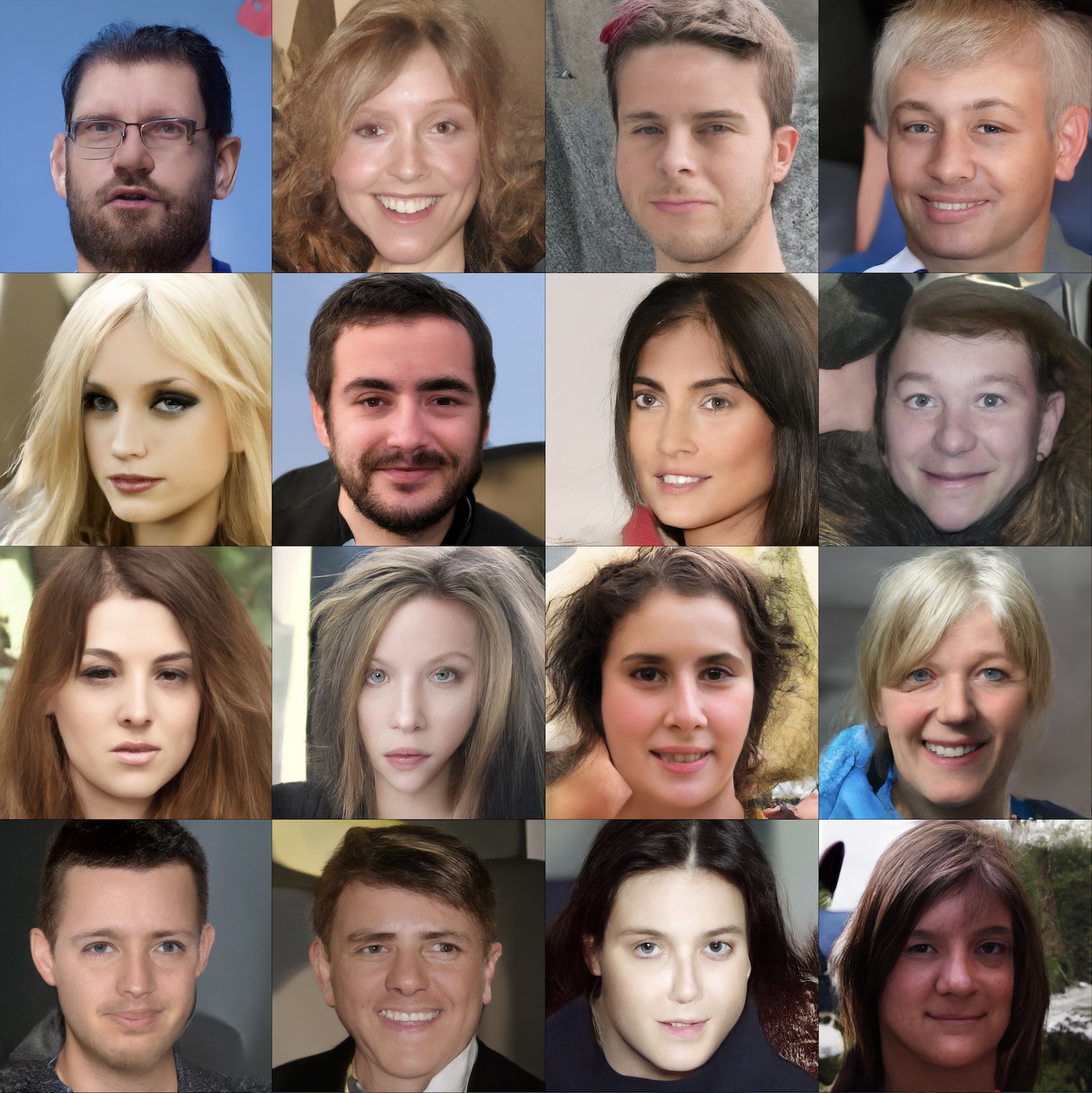}
    \caption{\textbf{Uncurated Visualization of FacesHQ dataset.}
    }
    \label{fig:ffhq1024_vis}
\end{figure*}

\noindent\textbf{FacesHQ $1024\times1024$ uncurated visualization} in~\cref{fig:ffhq1024_vis}.

\begin{figure*}
    \centering
\includegraphics[width=0.99\textwidth]{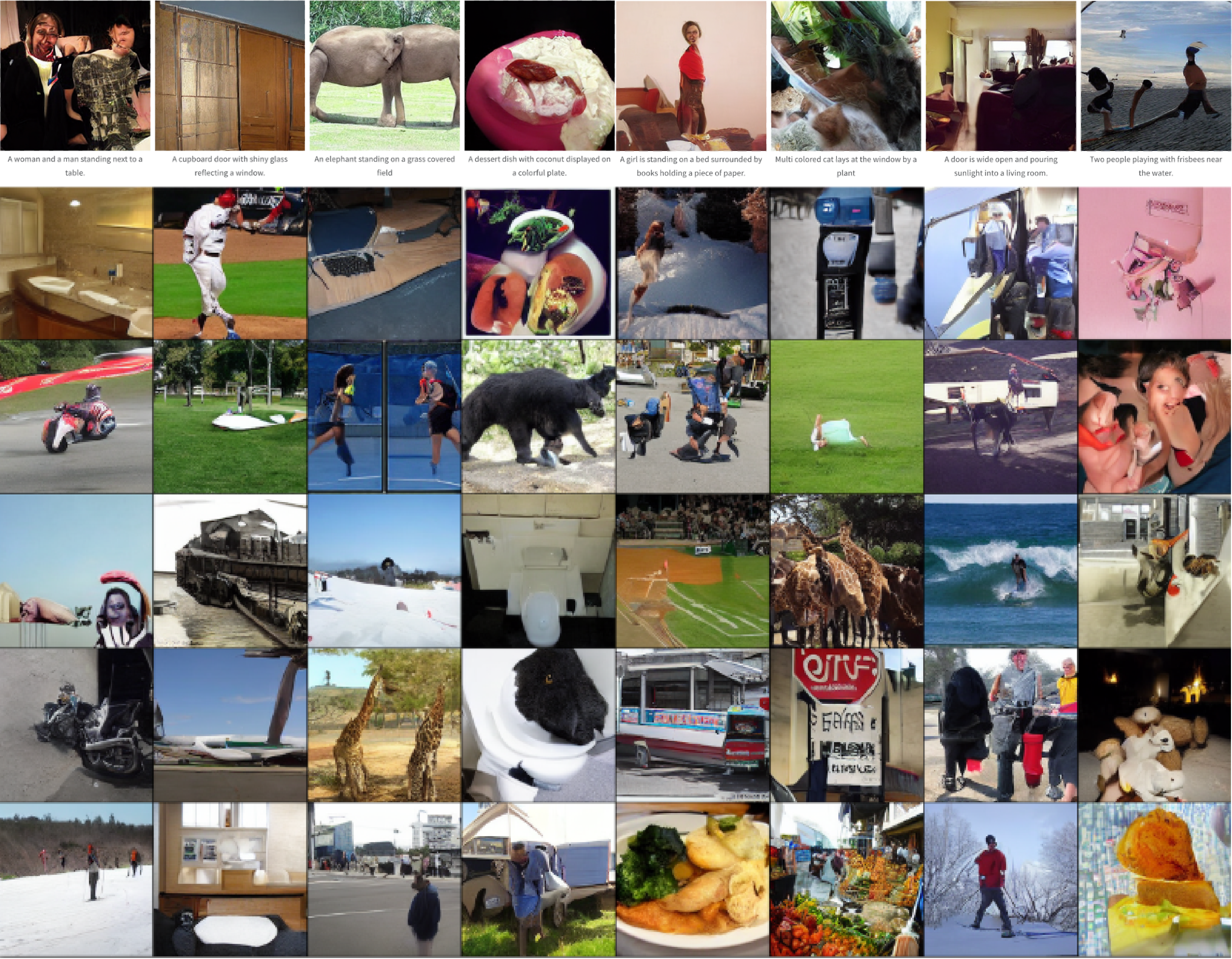}
    \caption{\textbf{The Uncurated Visualization of MS-COCO dataset.} The first row is illustrated with pairs of images and their captions, while the remaining rows only images.
    }
    \label{fig:coco_vis}
\end{figure*}

\noindent\textbf{MS-COCO uncurated visualization.} We visualize the samples in~\cref{fig:coco_vis}.

\begin{figure}
    \centering
    \begin{subfigure}{0.48\textwidth}
        \includegraphics[width=\textwidth]{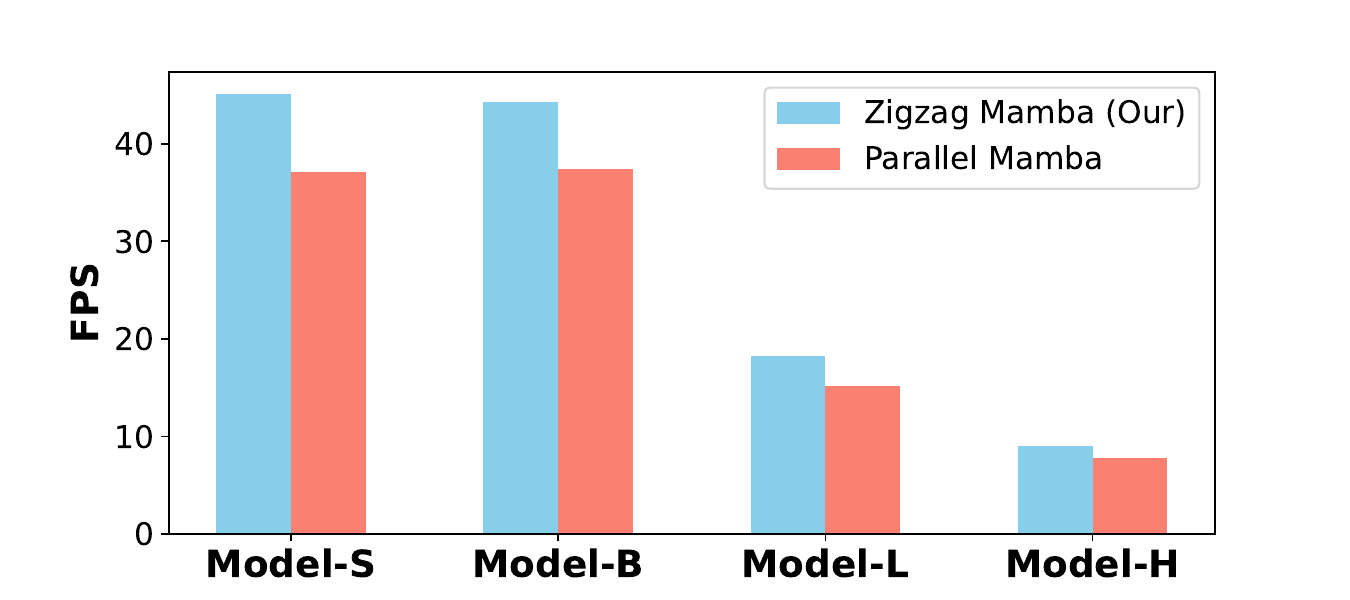}
       \caption{Model Complexity \textit{v.s.} FPS.}
    \end{subfigure}
    \hfill
    \begin{subfigure}{0.48\textwidth}
        \includegraphics[width=\textwidth]{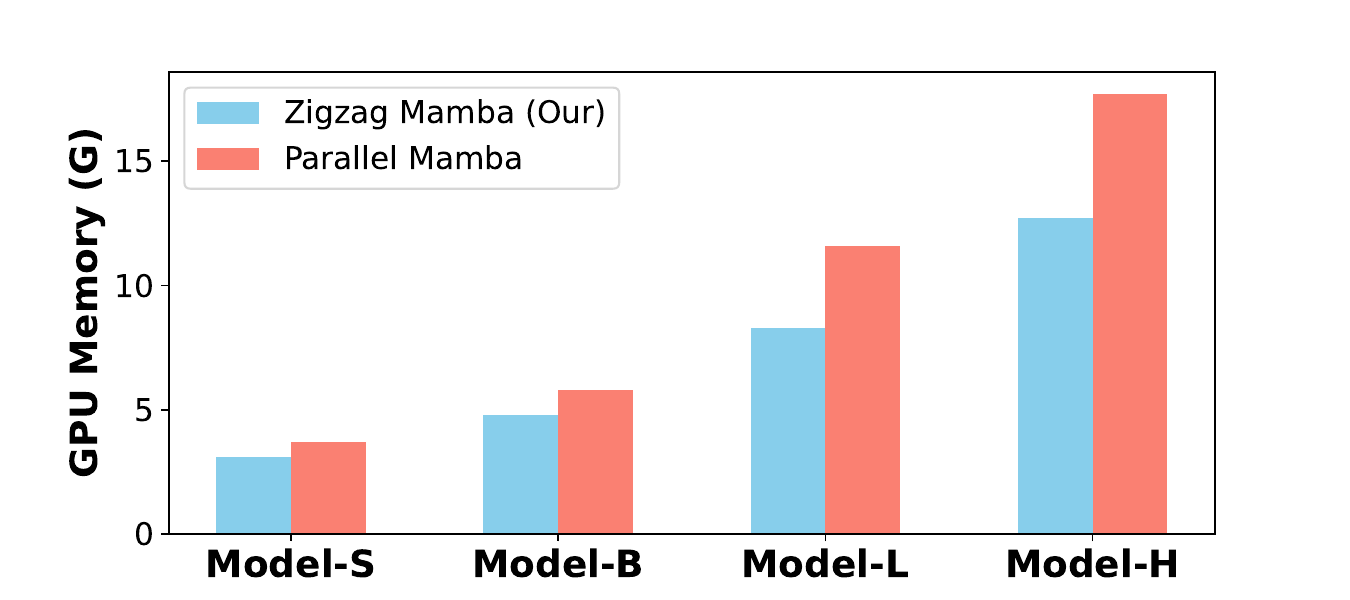}
       \caption{Model Complexity \textit{v.s.} GPU Memory.}
    \end{subfigure}
    \caption{\textbf{The ablation study about Model Complexity, FPS, GPU Memory}.
    }
     \label{fig:gpumem_fps_complexity_ablation}
\end{figure}

\begin{figure*}
    \centering
\includegraphics[width=0.95\textwidth]{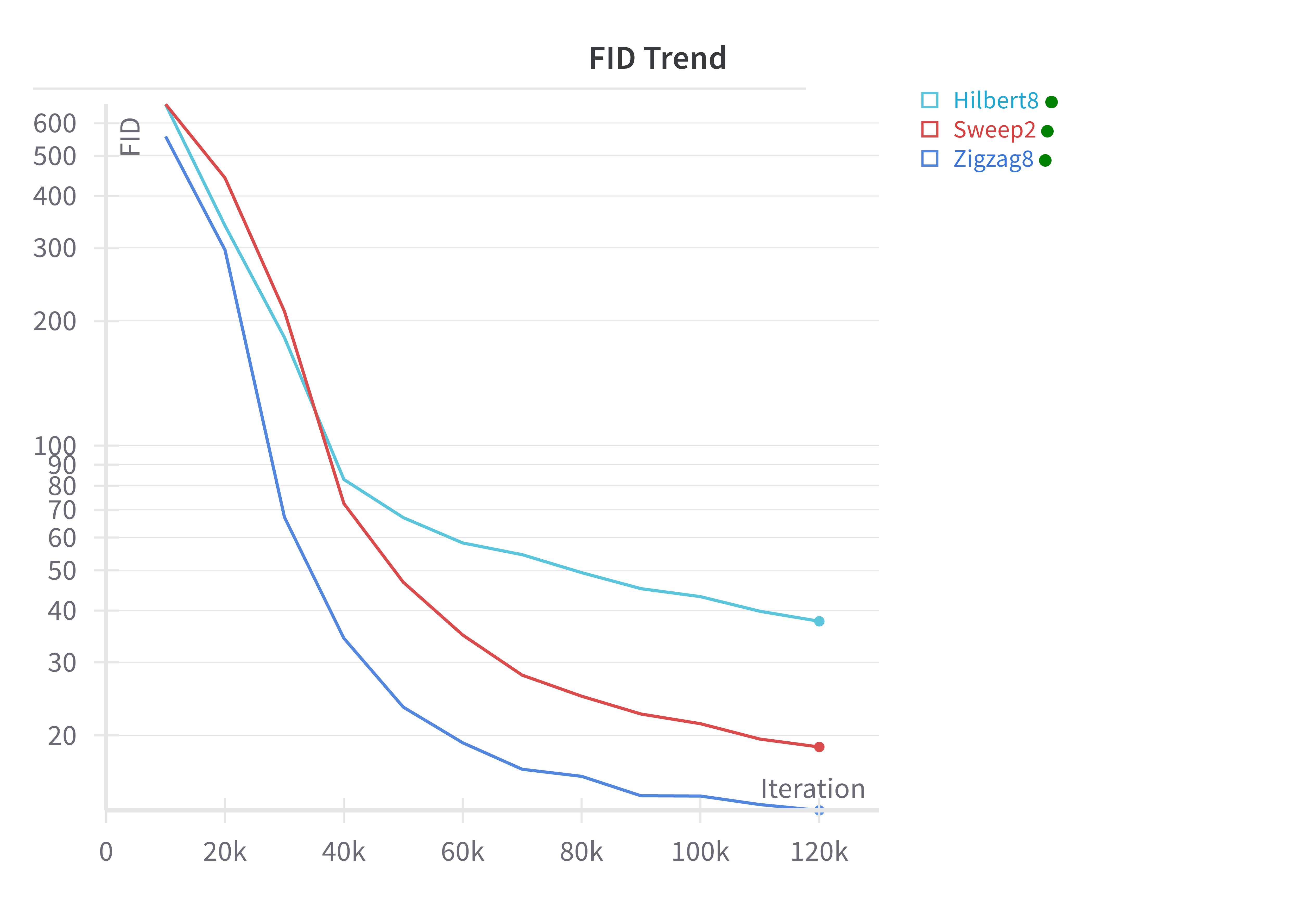}\hspace*{-2cm}
    \caption{\textbf{The FID trends comparing the Hilbert scan, Sweep scan, and our Zigzag scan.} The y-axis is   logarithmic scale.
    }
    \label{fig:hilbert_fid_trend}
\end{figure*}

\begin{figure}
    \centering
    \begin{subfigure}{0.48\textwidth}
        \includegraphics[width=\textwidth]{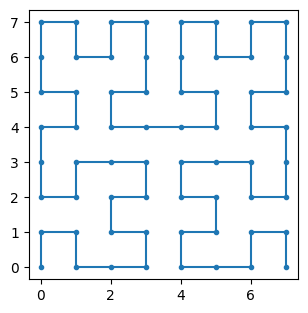}
       \caption{size=8}
    \end{subfigure}
    \hfill
    \begin{subfigure}{0.48\textwidth}
        \includegraphics[width=\textwidth]{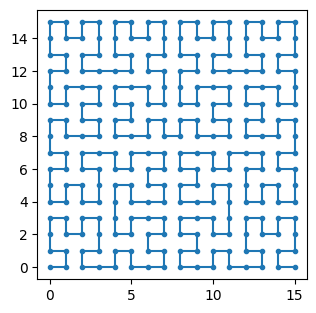}
       \caption{size=16}
    \end{subfigure}
    \caption{\textbf{The Hilbert space-filling curve with various sizes}.
    }
     \label{fig:hilbert_scan}
\end{figure}

\begin{figure*}
    \centering
\includegraphics[width=0.99\textwidth]{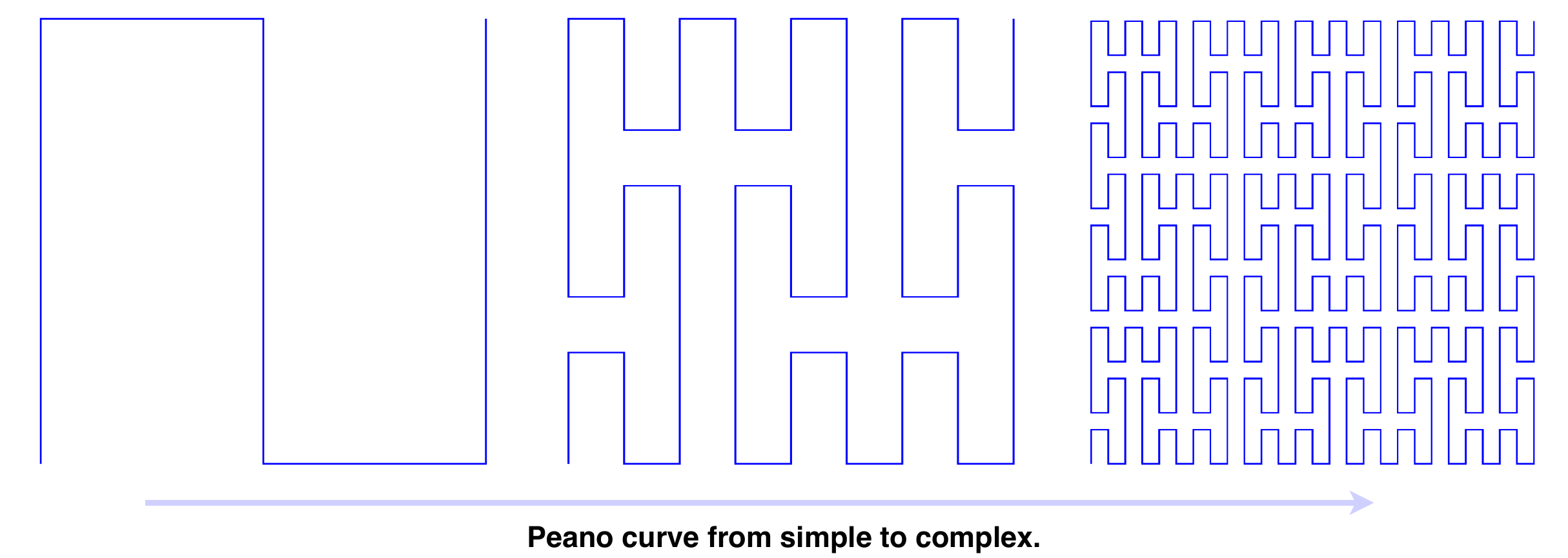}
    \caption{\textbf{The demonstration of Peano Curve.} The figure is borrowed from \url{https://en.wikipedia.org/wiki/Peano_curve}.
    }
    \label{fig:peano_curve_vis}
\end{figure*}

\subsection{Spatial Continuity is Critical}

\begin{figure}
    \centering
    \includegraphics[width=.6\textwidth]{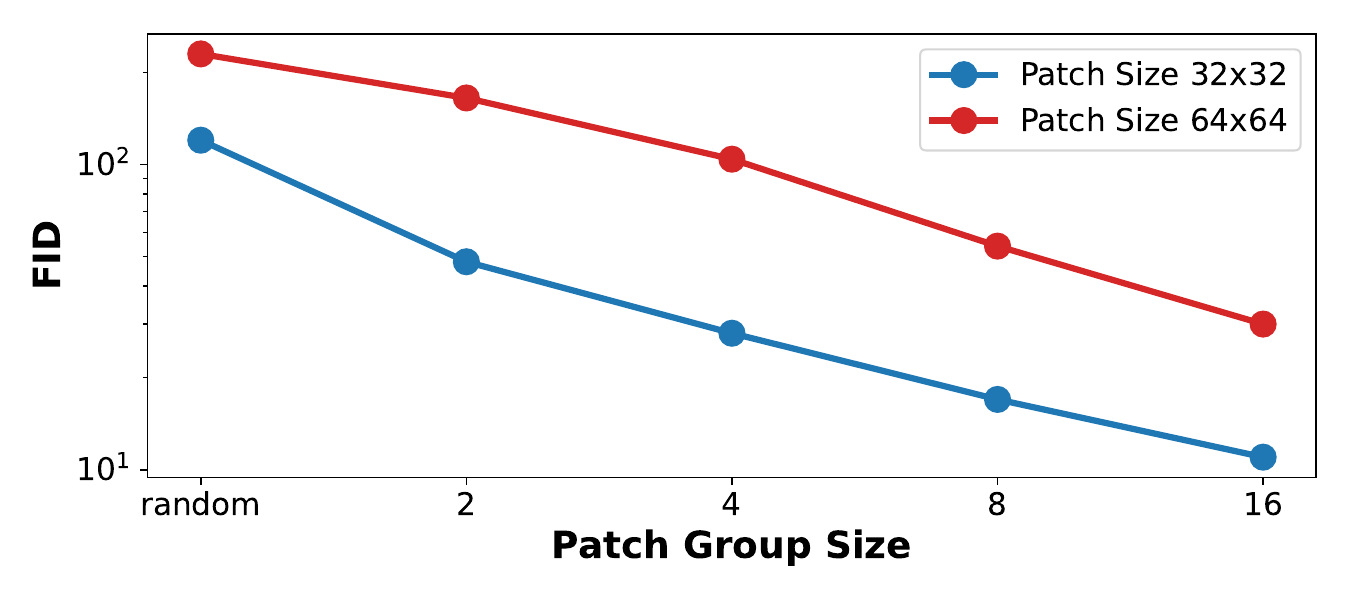}
    \caption{\textbf{Spatial Continuity Analysis.} As we incrementally enlarge the patch group size, the continuous segment of the patch also expands. This enhances spatial continuity, which we find improves FID on MultiModal-CelebA 256, 512 dataset.
    }
    \label{fig:motivation}
\end{figure}

We first explore the importance of spatial continuity in Mamba design by grouping patches of size $N \times N$ into various sizes: $2\times 2$, $4\times 4$, $8\times 8$, and $16\times 16$, resulting in groups of patch sizes $N/2 \times N/2$, $N/4 \times N/4$, $N/8 \times N/8$, and $N/16 \times N/16$, respectively. Then, we apply our designed Zigzag-8 scheme at the group level instead of the patch level. Figure~\ref{fig:motivation} illustrates that with increased spatial continuity, notably improved performance is achieved. Furthermore, we compare our approach with random shuffling of $N \times N$ patches, revealing notably inferior performance under random shuffling conditions. All of these results collectively indicate that spatial continuity is a critical requirement when applying Mamba in 2D sequences.

\subsection{Visualization}
\begin{figure}
    \centering
    \includegraphics[width=0.85\textwidth]{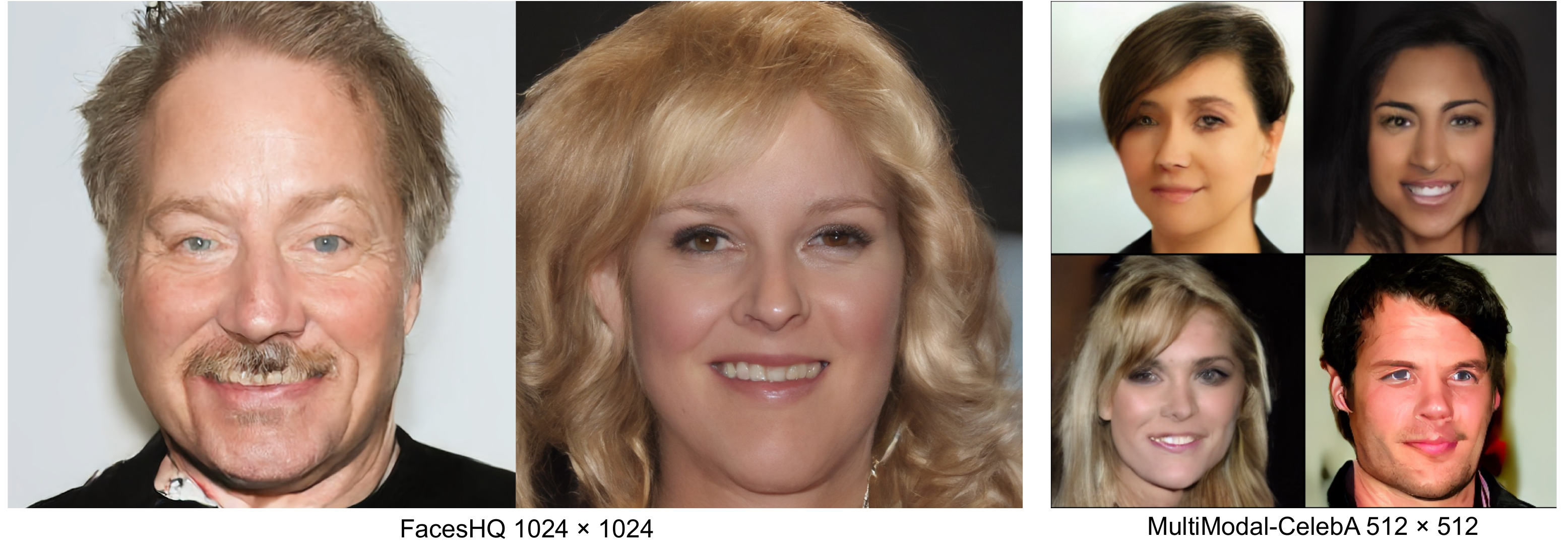}
    \caption{\textbf{Visualization of various resolutions on FacesHQ $1024\times1024$ and MultiModal-CelebA $512\times 512$.} Our generated samples present high fidelity across various resolutions.
   }
    \label{fig:vis_img}
\end{figure}

We demonstrate the image visualization of our best results on FacesHQ 1024 and MultiModal-CelebA 512 in Figure~\ref{fig:vis_img}. For the visualization of videos, please refer to Appendix~\ref{supp:sec:vis}. It is evident that the visualization is visually pleasing across various resolutions, indicating the efficacy of our methods.

\subsection{New Result about the Scanning Scheme}

We also conduct basic ablations on various factors, including position embedding and various Hilbert space-filling curves. Unlike the experiments in the main paper, we perform these experiments on unconditional MultiModal-CelebA256 dataset for a uniform comparison. We train the network for 100{,}000 steps.

\noindent\textbf{Exploration of Hilbert space-filling curve.} 
Primarily, we ablate the Hilbert scan curve~\cite{mckenna2019hilbert}, as depicted in ~\Cref{fig:hilbert_scan}. There are also eight variants of this scan considering different angles and starting points. We rearrange them in a similar manner to our Zigzag scan. All parameters are kept consistent for a fair comparison. We utilize the Gilbert algorithm~\footnote{https://github.com/jakubcerveny/gilbert} to guarantee that the Hilbert curve remains continuous across any square size.
We train our network on single A100-SXM4-80GB for 120k iterations. We evaluate the FID on 5{,}000 images for a fixed step, the FID curve is demonstrated in~\Cref{fig:hilbert_fid_trend}. 

While the Hilbert space-filling curve offers increased locality compared to our zigzag scan and maintains continuity, its complex structure appears to hinder the SSM's ability to work on the flattened sequence, resulting in a worse inductive bias than our zigzag curve on natural images. Therefore, we hypothesize that structure may hold greater significance than locality in generative tasks.

\paragraph{Hilbert Curve is difficult to optimize.} We show the result in ~\Cref{tab:ablate_hilbert}. We can observe that the performance of the Hilbert scan path drops significantly, even if we decrease the Order of Receptive Field (ORF). This confirms the assumption that the Hilbert scan path is difficult to optimize, even when considering only two different schemes of the Hilbert scan.

\paragraph{Another Interpretation: Zigzag scan is the simplest Peano curve.} Our Zigzag scan can be seen as the simplest case of Peano Curve as shown in~\Cref{fig:peano_curve_vis}.

\subsection{New Result of 2D visual data}
\label{supp:sec:moreresults}

\noindent\textbf{The variants of our ZigMa Models.} We list the variants of  our model in~\Cref{tab:models}. We use the Base (B) Model as the default. Applying the cross-attention model is optional, as this module can introduce some parameter and speed burdens. However, any advancements in attention optimization can be seamlessly integrated into our model.

\begin{figure}
    \centering
    \includegraphics[width=\textwidth]{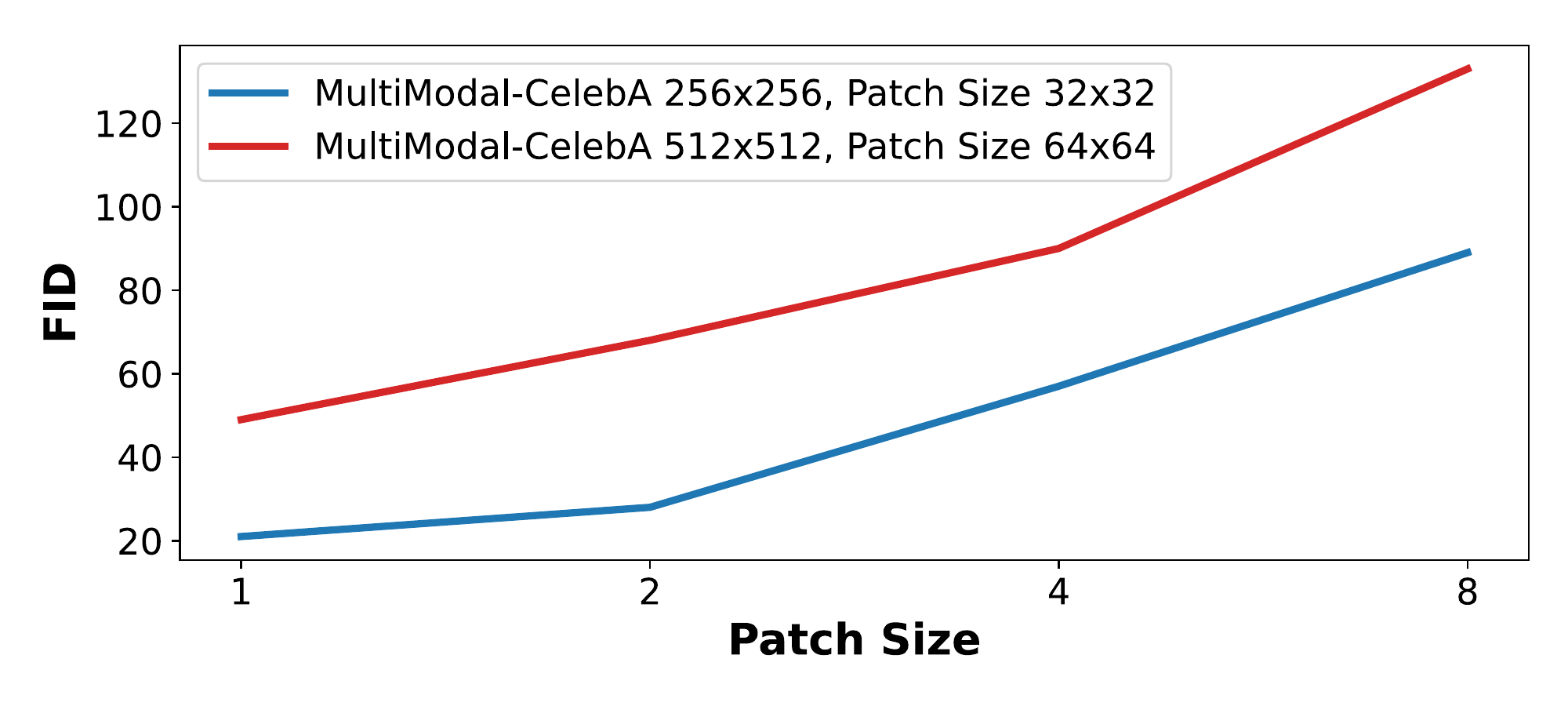}
       \caption{FPS \textit{v.s.} Patch Size.}
    \label{fig:patchsize_ablation}
\end{figure}

\noindent\textbf{Ablation of patch size.} We conducted an ablation study on patch sizes ranging from 1, 2, 4, to 8 in Figure~\ref{fig:patchsize_ablation}, aiming to explore their behaviors under the framework of Mamba. The results reveal that the FID deteriorates as the patch size increases, aligning with the common understanding observed in the field of transformers~\cite{dosovitskiy2020image_vit,transformer}. This suggests that smaller patch sizes are crucial for optimal performance.

\noindent\textbf{Ablation study about the Model Complexity and  FPS/GPU-Memory.} As shown in~\Cref{fig:gpumem_fps_complexity_ablation}. Our method can achieve much better parameter efficiency when incorporating the receptive order. The receptive order refers to the cumulative spatial-continuous zigzag scan path in 2D images, which we've incorporated into the Mamba as an inductive bias. We list the parameter consumption when we gradually increase the receptive order in~\Cref{fig:gpumem_fps_complexity_ablation}.   The receptive order refers to the cumulative spatial-continuous zigzag scan path in 2D images, which we've incorporated into the Mamba as an inductive bias.

\begin{figure}
    \centering
    \begin{subfigure}{0.48\textwidth}
        \includegraphics[width=\textwidth]{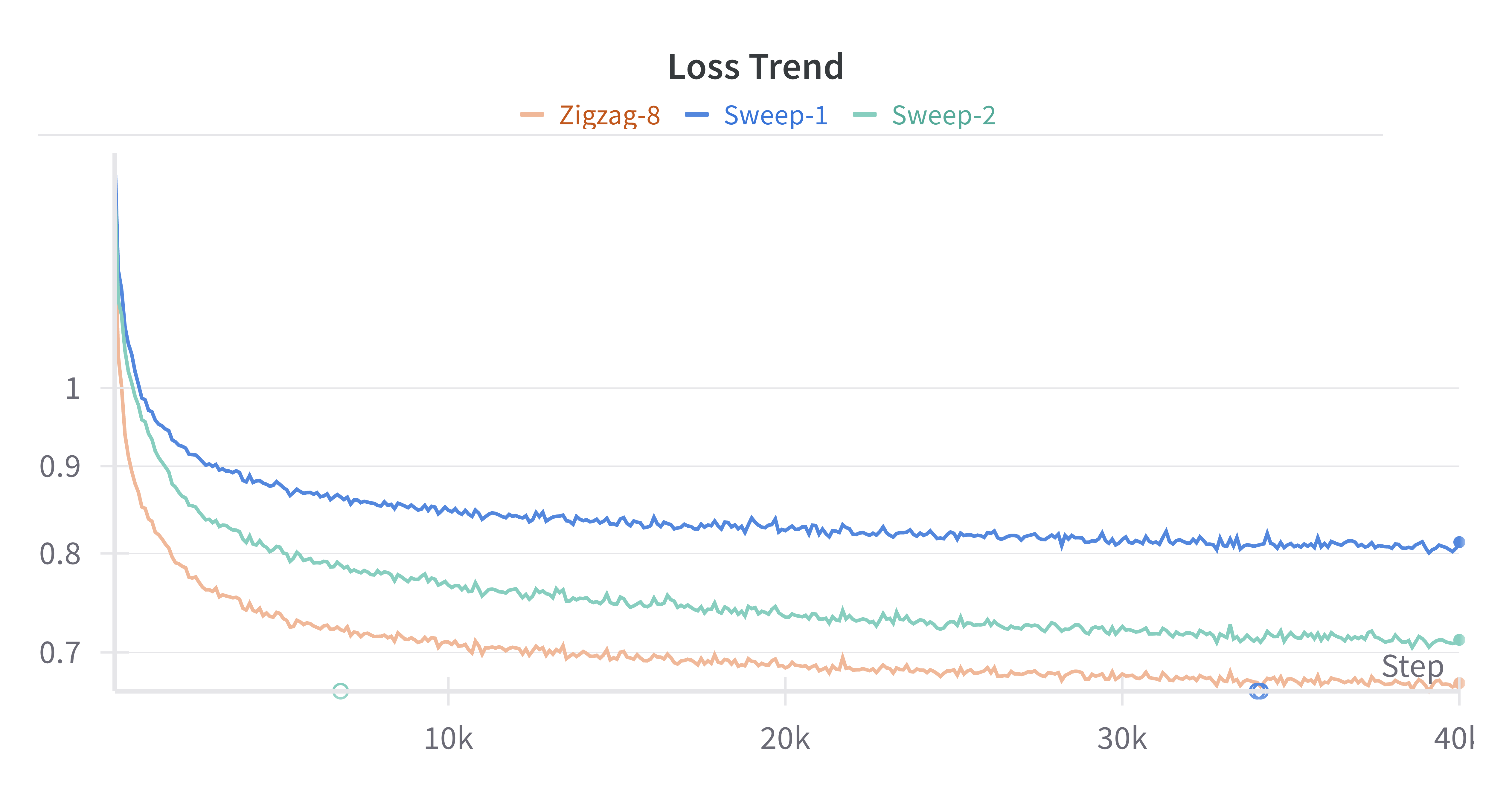}
       \caption{Loss trend of the MultiModal-CelebA256.}
    \end{subfigure}
    \hfill
    \begin{subfigure}{0.48\textwidth}
        \includegraphics[width=\textwidth]{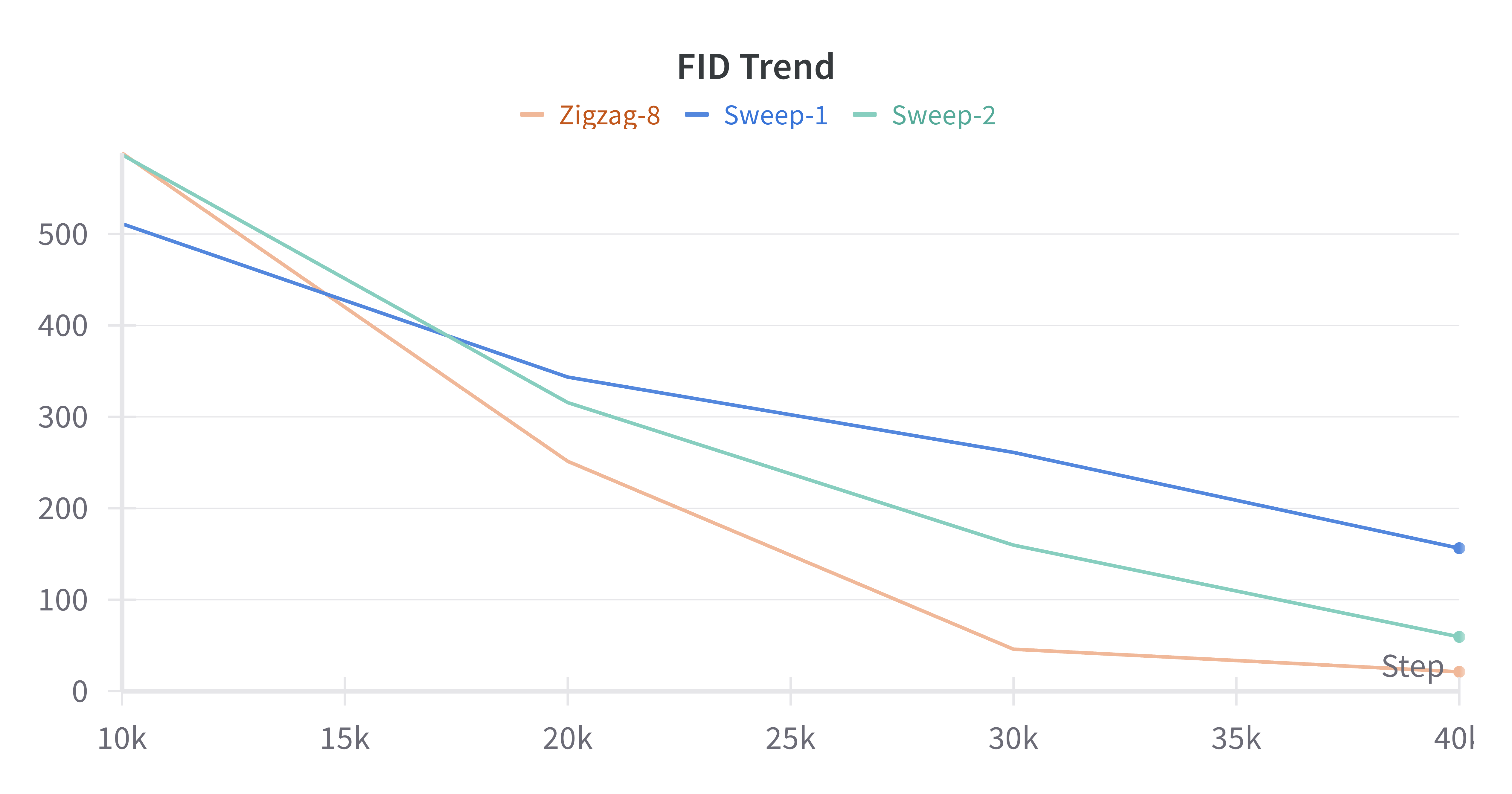}
       \caption{FID trend of the MultiModal-CelebA256.}
    \end{subfigure}
    \vskip\baselineskip
    \begin{subfigure}{0.48\textwidth}
        \includegraphics[trim=0 0 0 30pt, clip, ,width=\textwidth]{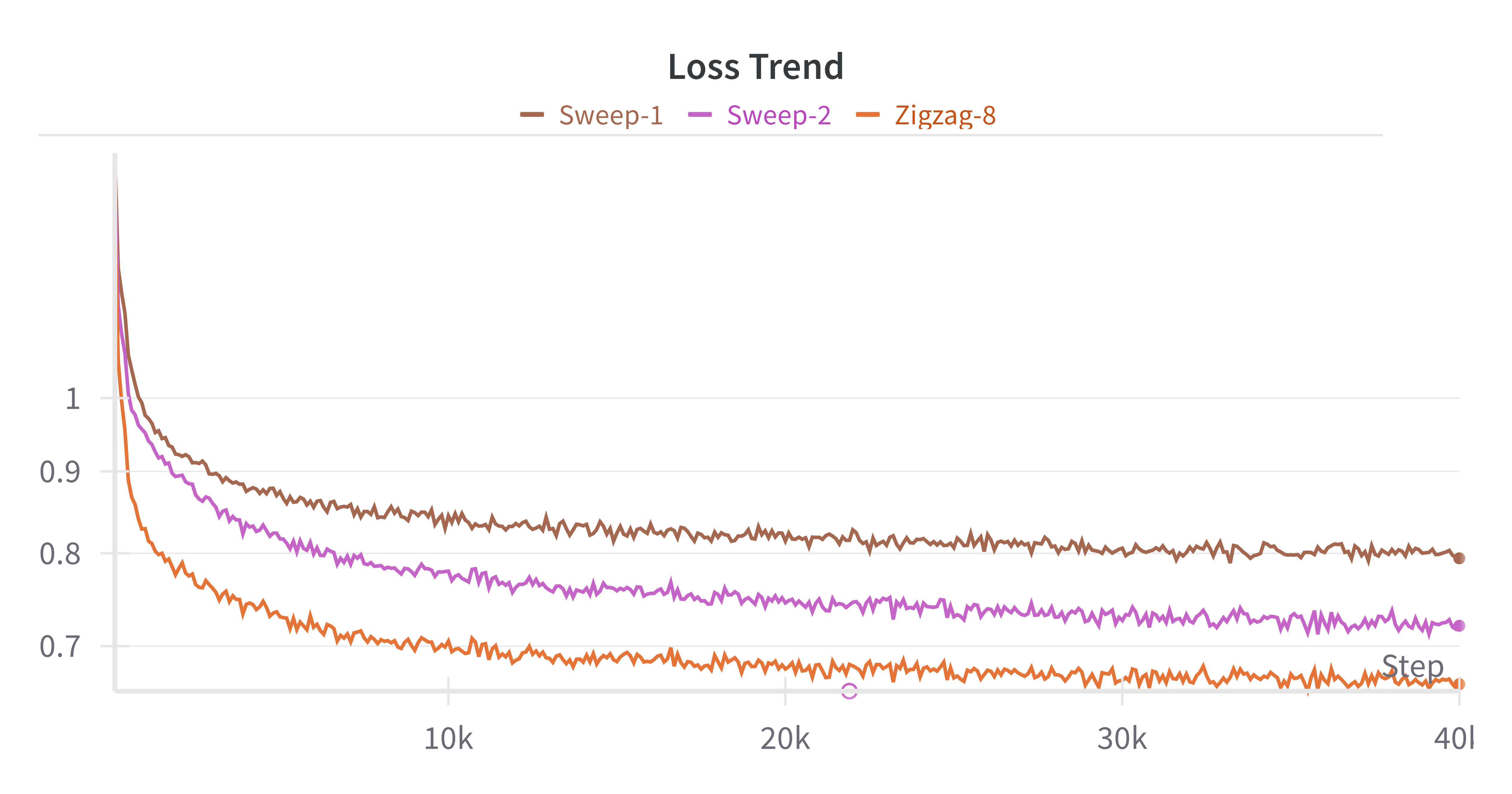}
       \caption{Loss trend of the MultiModal-CelebA512.}
    \end{subfigure}
    \hfill
    \begin{subfigure}{0.48\textwidth}
        \includegraphics[width=\textwidth]{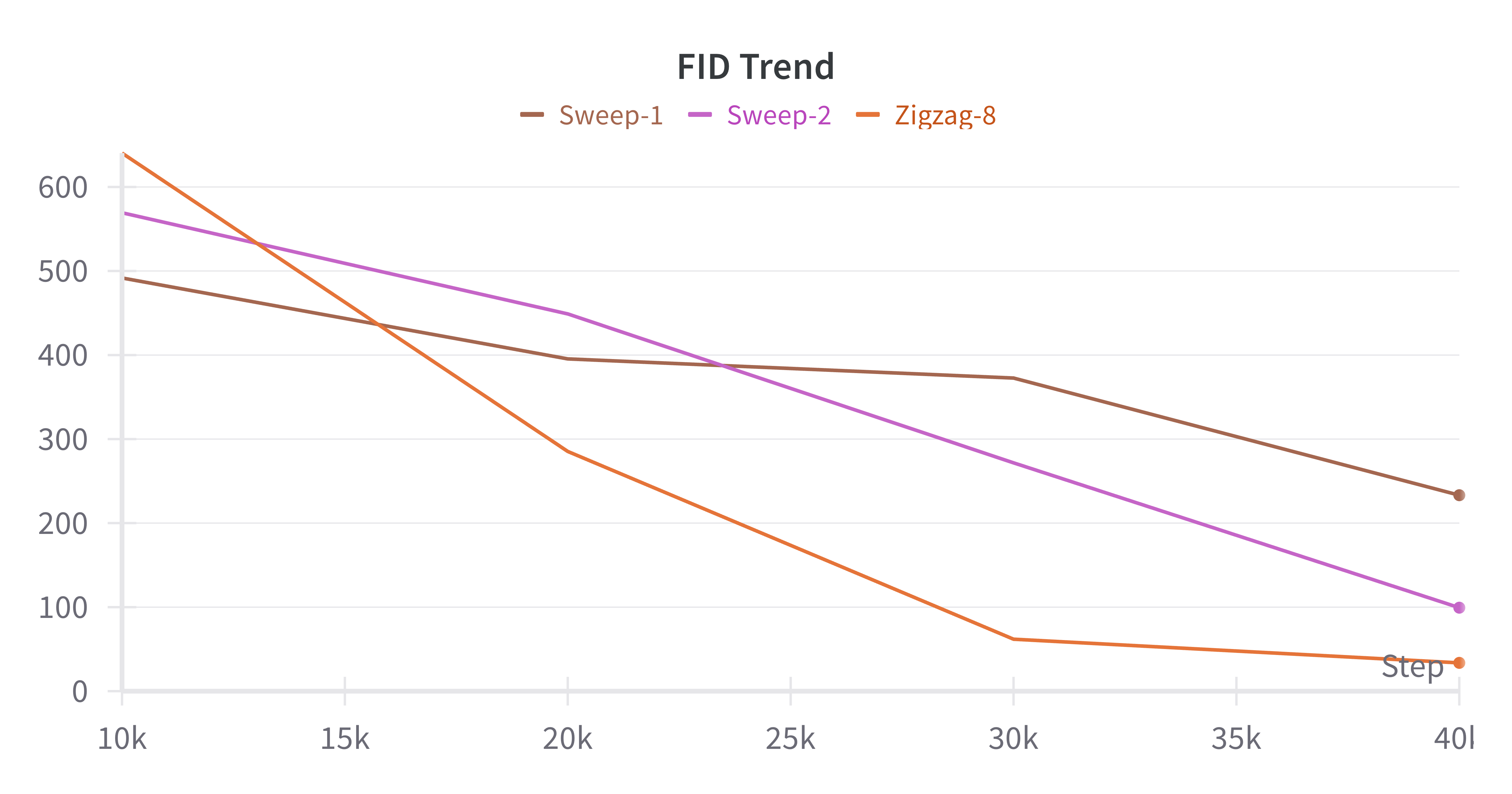}
       \caption{FID trend of the MultiModal-CelebA512.}
    \end{subfigure}
    \caption{\textbf{The loss and FID trend} under various resolutions on dataset MultiModal-CelebA. Sweep-1 and Sweep-2 are the Mamba scans without spatial continuity, while Zigzag-8 represents our method.
    This is the direct log from weight-and-bias (wandb).
    }
     \label{fig:loss_fid_trend}
\end{figure}

\noindent\textbf{Loss and FID curve.} The training loss curve and the FID curve are demonstrated in~\Cref{fig:loss_fid_trend}. The loss and FID show the same trend, with our Zigzag Mamba consistently outperforming other baselines like Sweep-1 and Sweep-2.

\noindent\textbf{In-context v.s. Cross Attention} We compare our cross-attention with in-context attention in~\Cref{tab:cross_attention}. For in-context attention, we concatenate the text tokens with the image tokens and feed them into the Mamba block. Our results demonstrate that in-context attention performs worse than our cross-attention. We hypothesize that this is due to the discontinuity between the text tokens and the image patch tokens. 
We discovered that PointMamba~\cite{liang2024pointmamba} arrives at the same conclusion and hypothesis as we do.

\subsection{New result of 3D Visual Data}

\noindent\textbf{The choice of the 3D Zigzag Mamba.} For Factorized 3D Zigzag Mamba in video processing, we deploy the \emph{sst} scheme for factorizing spatial and temporal modeling. This scheme prioritizes spatial information (ss)complexity over temporal information (t), hypothesizing that redundancy exists in the temporal domain. There are numerous other possible combinations of s and t to explore, which we leave for future work.

\subsection{More related works}

Several works~\cite{wang2024state,wang2023stablessm} have demonstrated that the State-Space Model possesses universal approximation ability under certain conditions. Mamba, as a new State-Space Model, has superior potential for modeling long sequences efficiently, which has been explored in various fields such as medical imaging~\cite{xing2024segmamba,umamba,wang2024mamba,ruan2024vm,gong2024nnmamba}, image restoration~\cite{guo2024mambair,zheng2024u_mamba_dehazing}, graphs~\cite{behrouz2024graphmamba,wang2024graph}, NLP word byte~\cite{wang2024mambabyte}, tabular data~\cite{ahamed2024mambatab}, human motion synthesis~\cite{motionmamba}, point clouds~\cite{liang2024pointmamba,zhang2024point},  image generation~\cite{fei2024scalable_dis}, semi-supervised learning~\cite{wang2024semi},interpretability~\cite{ali2024hidden}, image dehazing~\cite{zheng2024u_mamba_dehazing} and pan sharpening~\cite{he2024pan}. It has been extended to Mixture of Experts~\cite{anthony2024blackmamba}, spectral space~\cite{agarwal2023spectral}, multi-dimension~\cite{li2024mamba_nd,nguyen2022s4nd,visionmamba,liu2024vmamba} and dense connection~\cite{he2024densemamba}.
Among them, the most related to us are VisionMamba\cite{visionmamba,liu2024vmamba}, S4ND~\cite{nguyen2022s4nd} and Mamba-ND~\cite{li2024mamba_nd}.  VisionMamba\cite{visionmamba,liu2024vmamba} uses a bidirectional SSM in discriminative tasks which incurs a high computational cost. Our method applies a simple alternative mamba diffusion in generative models.
S4ND~\cite{nguyen2022s4nd} introduces local convolution into Mamba's reasoning process, moving beyond the use of only 1D data.  Mamba-ND~\cite{li2024mamba_nd} takes multi-dimensionality into account in discriminative tasks, making use of various scans within a single block. In contrast, our focus is on distributing scan complexity across every layer of the network, thus maximizing the incorporation of inductive bias from visual data with zero parameter burden.

Certain studies, such as Li's work in 2024~\cite{li2024denoising}, often explore the order of patches in token-based networks. However, while these studies concentrate on auto-regressive transformers, our focus is on the Mamba-based structure.

Several previous works~\cite{huang2021shuffle,zhang2018shufflenet} have focused on the shuffling operation to exchange information along the spatial or channel dimension. For instance, the Shuffle Transformer~\cite{huang2021shuffle} applies shuffling to spatial tokens to encourage cross-reasoning outside the attention windows. Our method follows the same approach. We shuffle the tokens to maintain a continuous spatial-filling scan path, promoting optimization across various layers. Given that the shuffling order differs across the layers, it could potentially avert the overfit problem~\cite{liu2024vmamba}.

\subsection{More Details}
\label{supp:details}

\noindent\textbf{Double-Indexing Issue for $\Omega_{i}$.} As shown in ~\cref{fig:framework}. We need to \textit{arrange} and  \textit{rearrange} operation that needs to conduct indexing along the token number dimension to achieve spatial-continuous mamba reasoning, as the indexing can be time-consuming~\footnote{We found that using the \texttt{torch.compile()} can largely ease the time issue, see \url{https://taohu.me/zigma} for more detail comparison.} when considering the large token numbers,  We can formulate the \texttt{arrange} and \texttt{rearrange} operation as follows:
\begin{align}
    {\Omega^{'}_{i}} &= \Bar{\Omega}_{i-1}\cdot \Omega_{i}, \\
    \bz_{i+1}  &= \texttt{scan}(\bz_{\Omega^{'}_{i}}),\\
\end{align}
where $\Bar{\Omega}_{-1}=I$, this assumes that the Mamba-based networks are permutation equivariant to the order of the tokens. They require 50\% fewer indexing operations, a point which we reiterate here for clearer comparison:

 \begin{align}
    \bz_{\Omega_{i}} &= \texttt{arrange}(\bz_i, \Omega_{i}), \\
    \Bar{\bz}_{\Omega_{i}} &= \texttt{scan}(\bz_{\Omega}) \label{eq:s6},\\
    \bz_{i+1} &= \texttt{arrange}(\Bar{\bz}_{\Omega_{i}},\Bar{\Omega_{i}}),
\end{align}

\noindent\textbf{Evaluation Metrics.} For image-level fidelity, we use established metrics such as Fréchet Inception Distance (FID) and Kernel Inception Distance (KID), following previous works. However, since studies~\cite{stein2024exposing,crowson2024scalable_hdit} have shown that FID does not fully reflect human-based opinions, we also adopt the Fréchet DINOv2 Distance (FDD) using the official repository~\footnote{https://github.com/layer6ai-labs/dgm-eval}. Our method primarily involves sampling 5,000 real and  5,000 fake images to compute the related metrics.

We primarily consider two metrics for video fidelity evaluation: framewise FID and Fréchet Video Distance (FVD)~\cite{unterthiner2019fvd}. We sample 200 videos and compute the respective metrics based on these samples.

We utilize the EMA models for evaluation, as they can deliver superior performance, as indicated in~\cite{yu2023video}.

\noindent\textbf{Extra Training Details.}  For text-conditioned generation, we conduct the experiments on the MultiModal-CelebA $256^2$,$512^2$~\cite{mmcelebahq} and MS COCO $256\times256$~\cite{coco} datasets. Both datasets are composed of text-image pairs for training. Typically, there are 5 to 10 captions per image in COCO and MultiModal-CelebA.    We convert discrete texts to a sequence of embeddings using a CLIP text encoder~\cite{radford2021learning_clip} following Stable Diffusion~\cite{rombach2022high_latentdiffusion_ldm}. Then these embeddings are fed into the network as a sequence of tokens.

\noindent\textbf{The training parameters of various datasets} are listed in~\cref{tab:architecture_default}.
We don't apply any position encoding because Mamba, unlike Transformer, is not permutation invariant. Therefore, its position is automatically encoded by its order in Mamba. Surprisingly, we also found that adding extra learnable position encoding can lead to better performance compared to the baseline. We hypothesize that these extra inductive biases can further benefit performance, even though the order of the tokens already incorporates some bias. 
For the COCO dataset, a weight decay of 0.01 can contribute to marginal FID gains (approximately 0.8).

\noindent\textbf{The conditioning of timestep and prompt.} The conditioning process is illustrated in~\cref{alg:mamba_block_crossattn}.  For the Mamba block, we incorporate the condition information. Specifically, we concatenate the condition token with the image patch token to enhance the conditioning mechanism.

\begin{table}
\setlength\tabcolsep{2.5pt}
\centering
\caption{\textbf{Hyperparameters and number of parameters for our network in various datasets.} 
} 
\scalebox{0.85}{
    \begin{tabular}{lcccc}
    \toprule
    \multicolumn{1}{c}{}  & {FacesHQ 1024}&{MS-COCO 256}&{MultiModal-CelebA 512}&{UCF-101}\\ \midrule
    Autoencoder $f$       & 8    &8         & 8 &8  \\
    $z$-shape             & $4 \times 128 \times 128$       &  $4 \times 32 \times 32$    &  $4 \times 64 \times 64$    &  $ 4 \times 32 \times 32$            \\
    Model size            & 133.8M  &133.8M & 133.8M & 133.8M      \\
    Patch size            & 2 & 1 & 1 & 2   \\
    Channels              & 768& 768 & 768 & 768  \\
    Depth                 & 12& 12& 12& 12 \\
    \hline
    Optimizer             & AdamW& AdamW& AdamW& AdamW\\
    Batch size/GPU            & 8& 8& 4& 8   \\
    GPU num            & 32& 32& 16& 16 \\
    Learning rate         & 1e-4 & 1e-4 & 1e-4 & 1e-4 \\
    weight decay & 0 &0&0&0\\
     EMA rate           & 0.9999& 0.9999& 0.9999& 0.9999 \\
      Warmup steps           & 0& 0& 0& 0 \\
    A100-hours&768&768&384&384\\    
    \bottomrule
    \end{tabular}
    }
\label{tab:architecture_default}
\end{table}

 \begin{algorithm}[H]
 \caption{%
 Mamba Block
 }
 \label{alg:mamba_block_crossattn}
 \definecolor{codeblue}{rgb}{0.25,0.5,0.5}
 \definecolor{codekw}{rgb}{0.85, 0.18, 0.50}
 \lstset{
   backgroundcolor=\color{white},
   basicstyle=\fontsize{9.2pt}{9.2pt}\ttfamily\selectfont,
   columns=fullflexible,
   breaklines=true,
   captionpos=b,
   commentstyle=\fontsize{9.2pt}{9.2pt}\color{codeblue},
   keywordstyle=\fontsize{9.2pt}{9.2pt}\color{codekw},
   escapechar={|}, 
   xleftmargin=.02\textwidth, xrightmargin=.02\textwidth
 }
 \begin{lstlisting}[language=python]
 def mamba_block(x, t, c = None):
       # x: input data, shape [B, (W x H) , C] or  [B, (T x W x H) , C]
       # t: timestep, (B, C)
       # c: condition, (B, D, C)       
       x = reshape(x) # (B, K, C)
       
       def _mamba(x):
           |\color{blue}x = rearrange(x)| # rearrange by a zigzag manner 
           x = mamba(x)
           |\color{blue}x = rearrange\_back(x)|# rearrange back by a zigzag manner 
        
       m, n = AdaLN( t )
       x = _mamba( x * m + n ) + x
           
       if c is not None:
           p, q = AdaLN( c )
           x = cross_attention( x * p + q ) + x
       return x 
 \end{lstlisting}
 \end{algorithm}

\begin{figure*}
    \centering
\includegraphics[width=0.99\textwidth]{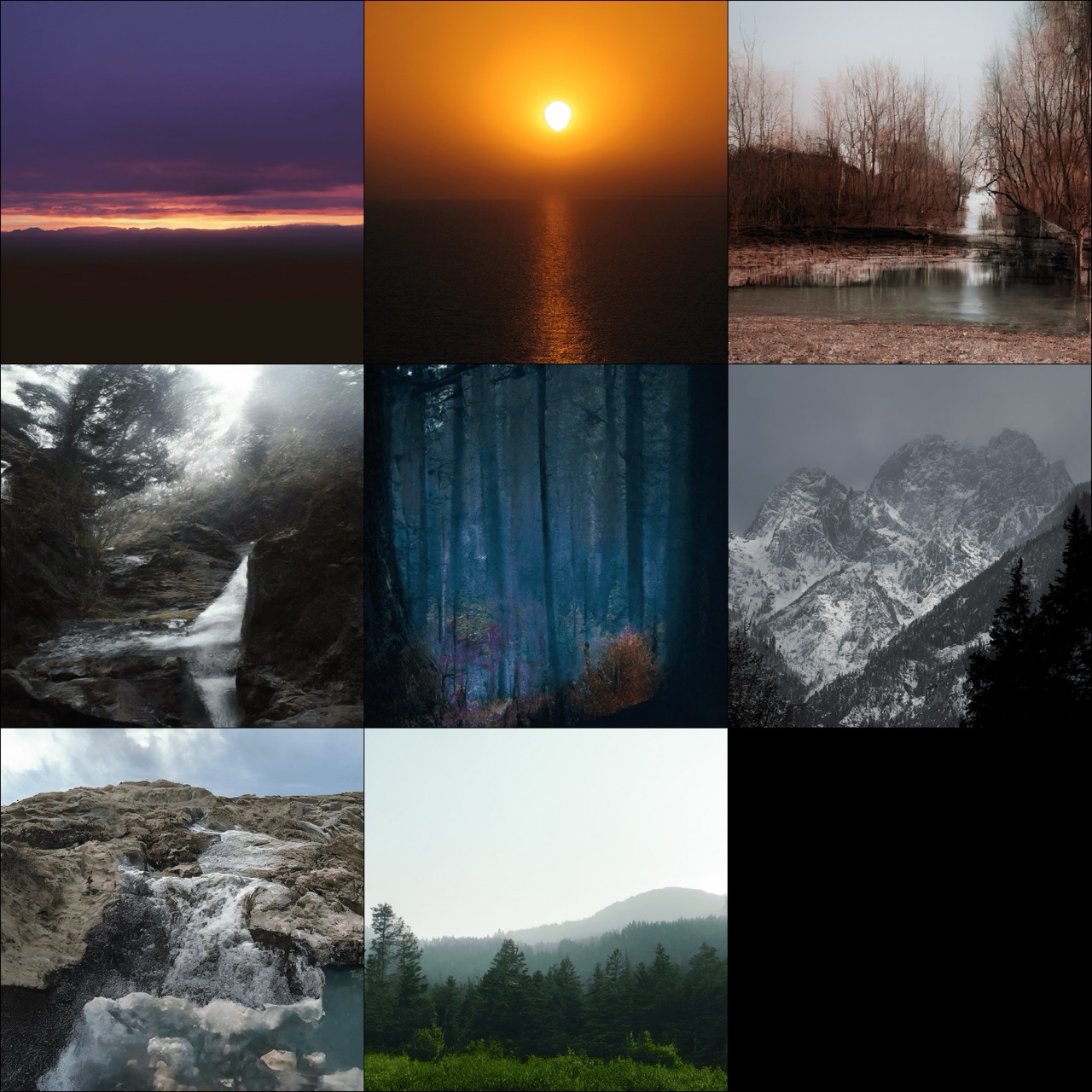}
    \caption{\textbf{Uncurated Visualization of Landscape HQ dataset~\cite{skorokhodov2021aligning_landscapehq}, with 5k FID of 10.07.}
    }
    \label{fig:landscapehq}
\end{figure*}

\end{document}